\documentclass[11pt,a4paper]{article}
\usepackage{amsfonts,amsmath,amssymb}
\usepackage{physics,mathtools,bm,mathrsfs,mleftright}
\usepackage{xcolor,enumerate,autobreak}
\usepackage{verbatim}
\allowdisplaybreaks[4]
\usepackage{wrapfig,multirow}
\usepackage{hyperref,prettyref}
\usepackage{natbib}
\usepackage{graphicx}
\usepackage{subcaption}
\usepackage{abstract}
\usepackage{tikz}

\newrefformat{figure}{Figure~\textup{\ref{#1}}}
\newrefformat{eq}{\textup{(\ref{#1})}}
\newrefformat{lem}{Lemma~\textup{\ref{#1}}}
\newrefformat{thm}{Theorem~\textup{\ref{#1}}}
\newrefformat{cor}{Corollary~\textup{\ref{#1}}}
\newrefformat{prop}{Proposition~\textup{\ref{#1}}}
\newrefformat{assu}{Assumption~\textup{\ref{#1}}}
\newrefformat{cond}{Condition~\textup{\ref{#1}}}
\newrefformat{remark}{Remark~\textup{\ref{#1}}}
\newrefformat{example}{Example~\textup{\ref{#1}}}
\newrefformat{def}{Definition~\textup{\ref{#1}}}
\newrefformat{sec}{Section~\textup{\ref{#1}}}
\newrefformat{subsec}{Subsection~\textup{\ref{#1}}}
\newrefformat{subsubsec}{Subsection~\textup{\ref{#1}}}

\newcommand{\ep}{\varepsilon}
\newcommand{\R}{\mathbb{R}}
\newcommand{\E}{\mathbb{E}}

\newcommand{\caB}{\mathcal{B}}

\newcommand{\caD}{\mathcal{D}}

\newcommand{\caI}{\mathcal{I}}

\newcommand{\caL}{\mathcal{L}}

\newcommand{\caN}{\mathcal{N}}

\newcommand{\caT}{\mathcal{T}}

\newcommand{\caX}{\mathcal{X}}

\newcommand{\bbP}{\mathbb{P}}
\newcommand{\bbE}{\mathbb{E}}
\newcommand{\bbR}{\mathbb{R}}

\newcommand{\mr}{\mathrm}
\providecommand{\ang}[1]{\left\langle{#1}\right\rangle}
\newcommand{\xk}[1]{\left(#1\right)}
\newcommand{\zk}[1]{\left[#1\right]}
\newcommand{\dk}[1]{\left\{#1\right\}}

\providecommand{\cref}{\prettyref}

\usepackage{amsthm}
\theoremstyle{plain}
\newtheorem{theorem}{Theorem}[section]
\newtheorem{lemma}[theorem]{Lemma}
\newtheorem{corollary}[theorem]{Corollary}
\newtheorem{condition}[theorem]{Condition}

\theoremstyle{definition}
\newtheorem{proposition}[theorem]{Proposition}

\newtheorem{remark}[theorem]{Remark}

\newtheorem{assumption}{Assumption}

\setcitestyle{numbers,square,comma,sort}
\usepackage{geometry,appendix}
\hypersetup{colorlinks=true,linkcolor=red,citecolor=green,urlcolor=blue}

\newcommand{\NN}{\mathrm{NN}}
\newcommand{\NTK}{\mathrm{NTK}}
\newcommand{\NT}{\mathrm{NT}}
\newcommand{\RF}{\mathrm{RF}}

\title{Neural Tangent Kernel of Neural Networks with Loss Informed by Differential Operators}
\author{Weiye Gan\thanks{
  \texttt{gwy22@mails.tsinghua.edu.cn}. Department of Mathematical Sciences, Tsinghua University, Beijing, China.
},~
Yicheng Li\thanks{
  \texttt{liyc22@mails.tsinghua.edu.cn}. Center for Statistical Science, Department of Industrial Engineering, Tsinghua University, Beijing, China.
},~
Qian Lin\thanks{\texttt{qianlin@tsinghua.edu.cn}. Center for Statistical Science, Department of Industrial Engineering, Tsinghua University, Beijing, China.
},~
Zuoqiang Shi\thanks{Corresponding author.
  \texttt{zqshi@tsinghua.edu.cn}. Yau Mathematical Sciences Center, Tsinghua University, Beijing, 100084, China \& Yanqi Lake Beijing Institute of Mathematical Sciences and Applications, Beijing, 101408, China.
}
}

\begin{document}
  \maketitle
  \begin{abstract}
      Spectral bias is a significant phenomenon in neural network training and can be explained by neural tangent kernel (NTK) theory. In this work, we develop the NTK theory for deep neural networks with physics-informed loss, providing insights into the convergence of NTK during initialization and training, and revealing its explicit structure. We find that, in most cases, the differential operators in the loss function do not induce a faster eigenvalue decay rate and stronger spectral bias. Some experimental results are also presented to verify the theory.
  \end{abstract}


  \section{Introduction}
  In recent years, Physics-Informed Neural Networks (PINNs)~\cite{raissi2019physics} are gaining popularity as a promising alternative to solve Partial Differential Equations (PDEs). PINNs leverage the universal approximation capabilities of neural networks to approximate solutions while incorporating physical laws directly into the loss function. This approach eliminates the need for discretization and can handle high-dimensional problems more efficiently than traditional methods~\cite{han2018solving,sirignano2018dgm}. Moreover, PINNs are mesh-free, making them particularly suitable for problems with irregular geometries~\cite{jagtap2020extended,yuan2022pinn}. Despite these advantages, PINNs are not without limitations. One major challenge is their difficulty in training, often resulting in slow convergence or suboptimal solutions~\cite{bonfanti2025challenges,krishnapriyan2021characterizing}. This issue is particularly pronounced in problems with the underlying PDE solutions that contain high-frequency or multiscale features~\cite{fuks2020limitations,raissi2018deep,zhu2019physics}.

To explain the obstacles in training PINNs, a significant aspect is about the deficiency of neural networks in learning multifrequency functions, referred to as spectral bias~\cite{rahaman2019spectral,xu2019training,geifman2022spectral,ronen2019convergence}, which means that neural networks tend to learn the components of "lower complexity" faster during training~\cite{rahaman2019spectral}. This phenomenon is intrinsically linked to the Neural Tangent Kernel (NTK) theory~\citep{jacot2018_NeuralTangent}, as the NTK's spectrum directly governs the convergence rates of different frequency components during training~\cite{cao2019towards}. Specifically,  neural networks are shown to converge faster in the directions defined by eigenfunctions of NTK with larger eigenvalues. Therefore, the components that are considered to have "low complexity" empirically are actually eigenfunctions of NTK with large eigenvalues and vice versa for components of high complexity. The detrimental effects of spectral bias can be exacerbated by two primary factors: first, the target function inherently possesses significant components of high complexity, and second, there is a substantial disparity in the magnitudes of the NTK's eigenvalues.

In the context of PINNs, the objective function corresponds to the solution of the PDE, rendering improvements in this aspect particularly challenging. A more promising avenue lies in refining the network architecture to ensure that the NTK exhibits a more favorable eigenvalue distribution. Several efforts have been made in this domain, such as the implementation of Fourier feature embedding~\cite{wang2021eigenvector,jin2024fourier} and strategic weight balancing to harmonize the disparate components of the loss function~\cite{wang2022and}. 

Different from the l2 loss in standard NTK theory, PINNs generally consider the following physics informed loss
\begin{equation*}
    \caL(u) = \frac{1}{2n}\sum_{i=1}^n(\caT u(x_i;\theta)-f_i)^2 + \frac{1}{2m}\sum_{j=1}^m(\caB u(x_j;\theta)-g_j)^2
\end{equation*}
for PDE
\begin{equation*}
      \label{eq:general_pde}
      \left\{\begin{array}{cc}
               \mathcal{D}u(x)=f(x), & x\in\Omega,         \\
               \mathcal{B}u(x)=g(x), & x\in\partial\Omega.
      \end{array}\right.
    \end{equation*}
In this paper, we only focus on the loss related to the interior euqation and neglect the boundary conditions, i.e.
\begin{equation*}
  \mathcal{L}(u;\mathcal{T})\coloneqq\frac{1}{2n}\sum_{i=1}^n(\mathcal{T}u(x_i;\theta)-f_i)^2.
\end{equation*}
This loss function may be adopted when $\Omega$ is a closed manifold or the boundary conditions are already hard constraints on neural networks~\cite{deng2024physical}. We first demonstrate the convergence of the neural network kernel at initialization. While most previous works only consider shallow networks. The idea based on functional analysis in \cite{hanin2021_RandomNeural} is applied so that we can process arbitrary high-order differential operator $\caT$ and deep neural networks. Another benefit of this approach is that we can show that the NTK related to $\caL(u;\caT)$ is exactly $\caT_x\caT_{x'}K^{NT}(x,x')$ where $K^{NT}(x,x')$ is the NTK for l2 loss $\caL(u;Id)$. With this connection, we analyze the impact of $\caT$ on the decay rate of the NTK's eigenvalues, which affects the convergence and generalization of the related kernel regression~\cite{li2024_EigenvalueDecay,li2024generalization}. We found that the additional differential operator in the loss function does not lead to a stronger spectral bias. Therefore, to improve the performance of PINNs from a spectral bias perspective, particular attention should be paid to the equilibrium among distinct components within the loss function~\cite{wang2022and,liu2024config}. For the convergence in training, we present a sufficient condition for general $\caT$ and neural networks. This condition is verified in a simple but specific case. Through these results, we hope to advance the theoretical foundations of PINNs and pave the way for their broader application in scientific computing.

The remainder of the paper is organized as follows. In \cref{sec:preliminary},  we introduce the function space we consider, the settings of neural networks, and some previous results of the NTK theory. All theoretical results are demonstrated in \cref{sec:results}, including the convergence of NTK during initialization and training and the impact of the differential operator within the loss on the spectrum of NTK. In \cref{sec:experiments}, we design some experiments to verify our theory.  Some conclusions are drawn in \cref{sec:conclusion}.

  \section{Preliminary}
  \label{sec:preliminary}
In this section, we introduce the basic problem setup including the function space considered and the neural network structure, as well as some background on the NTK theory.
\subsection{Continuously Differential Function Spaces}
Let $T\subset\R^n$ be a compact set, $k\geq 0$ be a integer and $Z_n$ be the $n$-fold index set,
\begin{equation*}
    Z_n=\{\alpha=(\alpha_1,\dots,\alpha_n)\big|\alpha_i \text{ is a non-negative integer, }\forall i=1,\dots,n\}.
\end{equation*}
We denote $\abs{\alpha}=\sum_{i=1}^n\alpha_i$ and $D^\alpha=\frac{\partial^{\abs{\alpha}}}{\partial x_1^{\alpha_1}\dots\partial x_n^{\alpha_n}}$.
Then, the $k$-times continuously differentiable function space $C^k(T)$ is defined as
\begin{equation*}
    C^k(T;\R^m)=\{u:T\rightarrow\R^m\big|D^\alpha u \text{ is continuous on } T,\ \forall\alpha:\abs{\alpha}\leq k\}.
\end{equation*}
If $m=1$, we disregard $\R^m$ for simplicty. The same applies to the following function spaces. $C^k(T;\R^m)$ can be equipped with norm
\begin{equation*}
    \norm{u}_{C^k(T;\R^m)}=\max_{\alpha\in Z_n,\abs{\alpha}\leq k}\sup_{x\in T}\abs{D^\alpha u(x)}.
\end{equation*}
For a contant $\beta\in[0,1]$, we also define
\begin{equation*}
    [u]_{C^{0,\beta}(T;\R^m)}=\sup_{x,y\in T,x\neq y}\frac{\abs{u(x)-u(y)}}{\abs{x-y}^\beta}
\end{equation*}
and
\begin{equation*}
    C^{0,\beta}(T;\R^m)=\{u\in C^0(T;\R^m)\big|[u]_{C^{0,\beta}(T;\R^m)}<\infty\},
\end{equation*}
\begin{equation*}
    C^{k,\beta}(T;\R^m)=\{u\in C^k(T;\R^m)\big|[D^\alpha u]_{C^{0,\beta}(T;\R^m)}<\infty,\ \forall \alpha:\abs{\alpha}\leq k\}.
\end{equation*}
$C^{0,1}(T;\R^m)$ is the familiar Lipschitz function space. $C^{k,\beta}(T;\R^m)$ can also be equipped with norm
\begin{equation*}
    \norm{u}_{C^{k,\beta}(T;\R^m)}=\norm{u}_{C^k(T;\R^m)}+\max_{\alpha\in Z_n,\abs{\alpha}\leq k}[D^\alpha u(x)]_{C^{0,\beta}(T;\R^m)}.
\end{equation*}
For a function of two variables $u(x,x')$, we denote $D_x^\alpha$ as a differential operator with respect to $x$, and similarly for $D_{x'}^\alpha$. We have analogous definitions,
\begin{equation*}
    C^{k\times k}(T\times T;\R^m)=\{u:T\times T\rightarrow\R^m\big|D_x^\alpha D_{x'}^{\alpha'} u \text{ is continuous on } T\times T,\ \forall\alpha,\alpha':\abs{\alpha},\abs{\alpha'}\leq k\}
\end{equation*}
with norm
\begin{equation*}
    \norm{u}_{C^{k\times k}(T\times T;\R^m)}=\max_{\alpha,\alpha'\in Z_n,\abs{\alpha},\abs{\alpha'}\leq k}\sup_{x,x'\in T}\abs{D_x^\alpha D_{x'}^{\alpha'} u(x,x')}
\end{equation*}
and
\begin{equation*}
    C^{k\times k,\beta}(T\times T;\R^m)=\{u\in C^{k\times k}(T\times T;\R^m)\big|[D_x^\alpha D_{x'}^{\alpha'} u]_{C^{0,\beta}(T\times T;\R^m)}<\infty,\ \forall\alpha,\alpha':\abs{\alpha},\abs{\alpha'}\leq k\}
\end{equation*}
with norm
\begin{equation*}
    \norm{u}_{C^{k\times k,\beta}(T\times T;\R^m)}=\norm{u}_{C^{k\times k}(T\times T;\R^m)}+\max_{\alpha,\alpha'\in Z_n,\abs{\alpha},\abs{\alpha'}\leq k}[D_x^\alpha D_{x'}^{\alpha'} u(x)]_{C^{0,\beta}(T\times T;\R^m)}.
\end{equation*}

With the chain rule and the fact that the composition of Lipschitz functions is still Lipschitz, the following lemmas can be verified.
\begin{lemma}
    \label{lem:fundamental-fact}
    Let $T_0\subset\R^{n_0}$, $T_1\subset\R^{n_1}$ be two compact sets. Let $\varphi:T_0\rightarrow T_1$ and $\psi:T_1\rightarrow \R^{n_2}$ be two $C^{k,1}$ maps. Then, $\psi\circ\varphi:T_0\rightarrow \R^{n_2}$ is also a $C^{k,1}$ map. And there exists a constant $C$ only depending on $k$, $\norm{\varphi}_{C^{k,1}(T_0;T_1)}$ and $\norm{\psi}_{C^{k,1}(T_1;\R^{n_2})}$ such that
    \begin{equation*}
        \norm{\psi\circ\varphi}_{C^{k,1}(T_0;\R^{n_2})}\leq C.
    \end{equation*}
\end{lemma}

\begin{lemma}
    \label{lem:fundamental-fact2}
    Let $T_0\subset\R^{n_0}$, $T_1\subset\R^{n_1}$ be two compact sets. Let $\varphi:T_0\times T_0\rightarrow T_1$ and $\psi:T_1\rightarrow \R^{n_2}$ be of class $C^{k\times k,1}$ and $C^{k,1}$ respectively. Then, $\psi\circ\varphi:T_0\times T_0\rightarrow \R^{n_2}$ is also a $C^{k\times k,1}$ map. And there exists a constant $C$ only depending on $k$, $\norm{\varphi}_{C^{k\times k,1}(T_0\times T_0;T_1)}$ and $\norm{\psi}_{C^{k,1}(T_1;\R^{n_2})}$ such that
    \begin{equation*}
        \norm{\psi\circ\varphi}_{C^{k\times k,1}(T_0\times T_0;\R^{n_2})}\leq C.
    \end{equation*}
\end{lemma}

\subsection{Settings of Neural Network}
Let the input $x \in \caX\subset\R^d$ where $\caX$ is a convex bounded domain and the output $y \in \R$. Let $m_0 = d$, $m_1,\dots,m_L$ be the width of $L$ hidden layers and $m_{L+1} = 1$.
Define the pre-activations $z^{l}(x) \in \R^{m_l}$ for $l = 1,\dots,L+1$ by
\begin{align}
  \label{eq:NN}
  \begin{aligned}
    &z^{1}(x) = W^{0}x,\\
    & z^{l+1}(x) = \frac{1}{\sqrt{m_l}}W^{l}\sigma(z^{l}(x)),\quad\text{for } l=1,\dots,L
  \end{aligned}
\end{align}
where $W^{l}\in \R^{m_{l+1} \times m_l}$ are the weights and $\sigma$ is the activation function satisfying the following assumption for some non-negative integer.
\begin{assumption}
  \label{assu:activation}
  There exists a nonnegative integer $k$ and positive constants $l_1,\dots,l_k$ such that activation $\sigma\in C^k(\bbR)$ and
  \begin{equation*}
    \norm{\frac{\sigma^{(j)}}{1+|x|^{l_j}}}_{L^\infty}<\infty
  \end{equation*}
  for all $j=1,\dots,k$.
\end{assumption}
The output of the neural network is then given by $u^{\NN}(x;\theta) = z^{L+1}(x)$.
Moreover, denoting by $z^{l}_i$ the $i$-th component of $z^l$, we have
\begin{equation*}
\begin{aligned}
  z^{1}_i(x) &= \sum_{j=1}^{m_0} W^{0}_{ij} x_j,\\
  z^{l+1}_i(x) &= \frac{1}{\sqrt{m_l}}\sum_{j=1}^{m_l} W^{l}_{ij} \sigma(z^{l}_j(x))
  \end{aligned}
\end{equation*}
We denote by $\theta = (W^{0},\dots,W^{L})$ the collection of all parameters flatten as a column vector. For simplicity, we also write $u(x) = u^{\NN}(x;\theta)$. The neural network is initialized by i.i.d random variables. Specifically, all elements of $W^l$ are i.i.d with mean $0$ and variance $1$.

In this paper, we consider to train neural network~\eqref{eq:NN} with gradient descent and the following physics informed loss
\begin{equation}
  \label{eq:loss}
  \mathcal{L}(u;\mathcal{T})\coloneqq\frac{1}{2n}\sum_{i=1}^n(\mathcal{T}u(x_i;\theta)-y_i)^2,
\end{equation}
where samples $x_i\in\caX$ and $\caT$ is a known linear differential operator.

\subsection{Training Dynamics}
When training neural network with loss~\eqref{eq:loss}, the gradient flow is given by
\begin{equation*}
  \dot{\theta} = - \nabla_{\theta} \mathcal{L}(u;\mathcal{T}) = - \frac{1}{n} \sum_{i=1}^n \nabla_{\theta} \mathcal{T} u(x_i;\theta) (\mathcal{T}u(x_i;\theta)-y_i).
\end{equation*}
Assuming that $u$ is sufficiently smooth, denoting $v = \caT u$, we have
\begin{equation*}
  \begin{aligned}
    \dot{v} &= \caT \dot{u}
    = \caT (\nabla_\theta u)^T \dot{\theta} = -  [\nabla_\theta (\caT u)]^T \nabla_{\theta} \mathcal{L}(u;\mathcal{T}) \\
    &=-\frac{1}{n}\sum_{i=1}^n[\nabla_\theta v]^T [\nabla_\theta v (x_i;\theta)] [v(x_i;\theta)-y_i].
  \end{aligned}
\end{equation*}
Define the time-varying neural network kernel (NNK)
\begin{equation}
  \label{eq:NNK}
  K_{\caT,\theta}(x,x') = \ang{\nabla_\theta (\caT u)(x;\theta),~\nabla_\theta (\caT u)(x';\theta)} =
  \ang{\nabla_\theta v(x;\theta),~\nabla_\theta v(x';\theta)}.
\end{equation}
Then the gradient flow of $v$ is just
\begin{equation}
  \label{eq:gradient-flow}
  \dot{v} = - \frac{1}{n} K_{\caT,\theta}(x,X) (v(X;\theta)-Y).
\end{equation}
NTK theory suggests that this training dynamic of $v$ is very similar to that of kernel regression when neural network is wide enough. And $K_{\caT,\theta}(x,x')$ is expected to converge to a time-invariant kernel $K_{\caT}(x,x')$ as width $m$ tends to infinity. If this assertion is true, we can consider the approximate kernel gradient flow of $v^{\NTK}$ by
\begin{equation}
  \label{eq:NTK_GradientFlow}
  \dot{v}^{\NTK}(x) = - \frac{1}{n} K_{\caT}^{\NT}(x,X) (v^{\NTK}(X;\theta)-Y),
\end{equation}
where the initialization $v^{\NTK}_0$ is not necessarily identically zero.
This gradient flow can be solved explicitly by
\begin{equation*}
  v^{\NTK}(x) = v^{\NTK}_0(x) + K_{\caT}^{\NT}(x,X)\varphi^{\mathrm{GF}}_t\left(\frac{1}{n}K_{\caT}(X,X)\right) (Y - v^{\NTK}_0(X)),
\end{equation*}
where $\varphi^{\mathrm{GF}}_t(z) \coloneqq (1-e^{-tz})/z$.

\subsection{NTK Theory}
When $\caT$ in~\eqref{eq:loss} is the identity map (denoted by $Id$), the training dynamic~\eqref{eq:gradient-flow} has been widely studied with the NTK theory~\citep{jacot2018_NeuralTangent}. This theory describes the evolution of neural networks during training in the infinite-width limit, providing insight into their convergence and generalization properties. It shows that the training dynamics of neural networks under gradient descent can be approximated by a kernel method defined by the inner product of the network's gradients. This theory has spurred extensive research, including studies on the convergence of neural networks with kernel dynamics~\citep{arora2019_ExactComputation,arora2019_FinegrainedAnalysis,du2018_GradientDescent,lee2019_WideNeural,allen-zhu2019_ConvergenceTheory}, the properties of the NTK~\citep{geifman2020_SimilarityLaplace,bietti2020_DeepEquals,li2024_EigenvalueDecay}, and its statistical performance~\citep{arora2019_FinegrainedAnalysis,hu2021_RegularizationMatters,lai2023_GeneralizationAbility}. By bridging the empirical behavior of neural networks with their theoretical foundations, the NTK provides a framework for understanding gradient descent dynamics in function space. In this section, we review some existing conclusions that are significant in deriving our results.
\paragraph{The random feature and neural tangent kernel}
For the finite-width neural network \eqref{eq:NN},
let us define the random feature kernel $K^{\RF,m}_l$ and the neural tangent kernel $K^{\NT,\theta}_l$ for $l=1,\dots,L+1$ by
\begin{equation*}
\begin{aligned}
  K^{\RF,m}_l(x,x') &= \mathrm{Cov}\left( z^{l}(x), z^{l}(x') \right),\\
  K^{\NT,\theta}_{l,ij}(x,x') &= \ang{\nabla_{\theta} z^{l}_i(x), \nabla_{\theta} z^{l}_j(x')},\qq{for} i,j=1,\dots,m_l.
  \end{aligned}
\end{equation*}
Note here that $K^{\RF,m}_l$ is deterministic and $K^{\NT,\theta}_l$ is random.
Since $m_{l+1} =1 $, we denote $K^{\NT,\theta}_{L+1}(x,x') = K^{\NT,\theta}_{L+1,11}(x,x')$.

Moreover, for the kernels associated with the infinite-width limit of the neural network, let us define
\begin{equation*}
  \begin{aligned}
    K^{\RF}_1(x,x') = K^{\NT}_1(x,x') = \ang{x,x'}.
  \end{aligned}
\end{equation*}
and the recurrence formula for $l=2,\dots,L+1$,
\begin{align}
  \label{eq:NTK_DefRecur}
  \begin{aligned}
    K^{\RF}_{l}(x,x') &= \E_{(u,v)\sim N(\bm{0},\bm{B}_{l-1}(x,x'))}\left[ \sigma(u) \sigma(v) \right], \\
    K^{\NT}_{l}(x,x') &= K^{\RF}_{l}(x,x') + K^{\NT}_{l-1}(x,x') \E_{(u,v)\sim N(\bm{0},\bm{B}_{l-1}(x,x'))}\left[ \sigma^{(1)}(u) \sigma^{(1)}(v) \right],
  \end{aligned}
\end{align}
where the matrix $\bm{B}_{l-1}(x,x') \in \R^{2 \times 2}$ is defined as:
\[\bm{B}_{l-1}(x,x') =
\begin{pmatrix}
  K^{\RF}_{l-1}(x,x)  & K^{\RF}_{l-1}(x,x')   \\
  K^{\RF}_{l-1}(x,x') & K^{\RF}_{l-1}(x',x').
\end{pmatrix}\]
The smoothness of the kernel $K_l^\RF$ and $K_l^\NT$ can be derived by the regularity of the activation function $\sigma$. The proof is presented in \cref{sec:smoothness}.
\begin{lemma}
    \label{lem:bound-diff-NTK}
    Let $k\geq 1$ be an integer and $\sigma$ satisfies \cref{assu:activation} for $k$. Then the kernel $K_l^\RF$ and $K_l^\NT$ defined as \eqref{eq:NTK_DefRecur} is of classes $C^{k\times k}$ and $C^{(k-1)\times(k-1)}$ respectively.
\end{lemma}

\paragraph{Convergnce of NNK}
The most basic and vital conclusion for NTK is that NNK defined as~\eqref{eq:NNK} converges to a time-invariant kernel when the width of the neural network tends to infinity.
\begin{lemma}[Proposition 34 in~\cite{li2024_EigenvalueDecay}]
\label{lem:C0-convergence}
  Consider NNK $K_{\caT,\theta}$ defined as \eqref{eq:NNK} where $\caT\equiv Id$ is the identity map. Let $\delta\in(0,1)$. Under proper assumptions on the parameters $\theta_t$, there exists constants $C_1>0$ and $C_2\geq 1$ such that, with probability at least $1-\delta$,
  \begin{equation*}
    \sup_{t\geq 0}\abs{ K_{Id,\theta_t}(z,z')-K_{Id}^{\NT}(x,x') } =O\left(m^{-\frac{1}{12}}\sqrt{\ln m} \right)
  \end{equation*}
  when $m\geq C_1\xk{\ln\xk{C_2/\delta}}^5$ and $\norm{z-x}_2$, $\norm{z'-x'}_2\leq O(1/m)$.
\end{lemma}

\paragraph{Convergence at initialization}
As the width tends to infinity, it has been demonstrated that the neural network converges to a Gaussian process at initialization.
\begin{lemma}[\cite{hanin2021_RandomNeural}]
  \label{lem:GaussianProcessAtInitialization}
  Fix a compact set $T \subseteq \mathbb{R}^{n_0}$.
  As the hidden layer width $m$ tends to infinity, the sequence of stochastic processes
  $x \mapsto u^{\NN}(x;\theta)$ converges weakly in $C^0(T)$ to a centered Gaussian process with covariance function
  $K^{\RF}_L$.
\end{lemma}






  \section{Main Results}
  \label{sec:results}
In this section, we present our main results. We first establish a general theorem concerning the convergence of NTKs at initialization. A sufficient condition for convergence in training is proposed and validated in some simple cases. Finally, leveraging the aforementioned results, we examine the impact of differential operators within the loss function on the spectral properties of the NTK.
\subsection{Convergence at initialization}
Let $S$ be a metric space and $X, X^n, ~n\geq 1$ be random variables taking values in $S$.
We recall that $X^n$ converges weakly to $X$ in $S$, denoted by $X^n \xrightarrow{w} X$, iff
\begin{align*}
  \E f(X^n) \to \E f(X) \qq{for any continuous bounded function} f: S \to \R.
\end{align*}
The first result is to show that if the activation function has a higher regularity, we can generalize \cref{lem:C0-convergence} to the case of weak convergence in $C^k$.
\begin{theorem}[Convergence of initial function]
  \label{thm:init_GP}
  Let $T\subset\caX$ be a compact set. $k\geq 0$ is an integer. $\sigma$ satisfies \cref{assu:activation} for $k$.
  For any $\alpha$ satisfying $|\alpha|\leq k$, fixed $l = 2,\dots,L+1$,
  fixing $m_{l}$, as $m_1,\dots,m_{l-1} \to \infty$,
  the random process
  \begin{align*}
    x \in \caX \mapsto D^\alpha z^{l}(x) \in \R^{m_l}
  \end{align*}
  converge weakly in $C^0(T;\R^{m_l})$ to a Gaussian process in $\R^{m_l}$ whose components are i.i.d. and have mean zero and covariance kernel $D_x^\alpha D_{x'}^\alpha K^{\RF}_l(x,x')$.
\end{theorem}

\begin{proof}
  With \cref{lem:tightness_Ck}, \cref{lem:Lipschitz-control-net} and Proposition 2.1 in \cite{hanin2021_RandomNeural}, we obtain the conclusion.
\end{proof}

With the aid of \cref{thm:init_GP}, the uniform convergence of NNK at initialization is demonstrated as follows.
\begin{theorem}
  \label{thm:init_NTK}
  Let $T\subset\caX$ be a convex compact set. $k\geq 0$ is an integer. $\sigma$ satisfies \cref{assu:activation} for $k+2$.
  For any $\alpha,\beta$ satisfying $|\alpha|,|\beta|\leq k$, fixed $l = 2,\dots,L+1$,
  fixing $m_{l}$, as $m_1,\dots,m_{l-1} \to \infty$ sequentially,
  we have
  \begin{align}
  \label{eq:kernel_convergence}
    D_x^\alpha D_{x'}^\beta K^{\NT,\theta}_{l,ij}(x,x') \xrightarrow{p} \delta_{ij} D_x^\alpha D_{x'}^\beta K^{\NT}_l(x,x')
    \qq{for} i,j=1,\dots,m_l
  \end{align}
  under $C^0(T\times T; \bbR).$
\end{theorem}

\begin{proof}
  The proof is completed by induction. When $l=1$, for any $i,j=1,\dots,m_1$,
  \begin{equation*}
    \begin{aligned}
      \quad K_{1,ij}^{\NT,\theta}(x,x')
      &=\ang{\nabla_{\theta}z_i^1(x),\nabla_\theta z_j^1(x')}\\
      &=\ang{\nabla_{\theta}W_i^0x,\nabla_\theta W_j^0x'}\\
      &=\delta_{ij}\ang{x,x'}\\
      &=\delta_{ij}K_{1}^{\NT}(x,x').
    \end{aligned}
  \end{equation*}
  Suppose that as $m_1,\dots,m_{l-1} \to \infty$, we have
  \begin{equation*}
    K^{\NT,\theta}_{l,ij}(x,x') \xrightarrow{p} \delta_{ij} K^{\NT}_l(x,x') \qq{under} C^{k\times k}(T\times T;\R)
    \qq{for} i,j=1,\dots,m_l.
  \end{equation*}
  For any $i,j=1,\dots,m_{l+1}$,
  \begin{equation}
  \label{eq:kernel}
    \begin{aligned}
      &\quad K_{l+1,ij}^{\NT,\theta}(x,x')\\
      &=\ang{\nabla_{\theta}z_i^{l+1}(x),\nabla_\theta z_j^{l+1}(x')}\\
      &=\delta_{ij}\frac{1}{m_l}\sum_{q=1}^{m_l}\sigma(z_q^l(x))\sigma(z_q^l(x'))\\
      &+\frac{1}{m_l}\sum_{q_1=1}^{m_l}\sum_{q_2=1}^{m_l}W_{iq_1}^lW_{jq_2}^l\sigma^{(1)}(z_{q_1}^l(x))\sigma^{(1)}(z_{q_2}^l(x'))K_{l,q_1q_2}^{\NT,\theta}(x,x').
    \end{aligned}
  \end{equation}
  With the induction hypothesis and \cref{thm:init_GP}, for any fixed $m_l$, as $m_1,\dots,m_{l-1} \to \infty$ sequentially,
  \begin{equation*}
    \begin{aligned}
      K_{l+1,ij}^{\NT,\theta}(x,x')&\xrightarrow{w}\frac{\delta_{ij}}{m_l}\sum_{q=1}^{m_l}\sigma(G_q^l(x))\sigma(G_q^l(x'))\\
      &+\frac{K^{\NT}_l(x,x')}{m_l}\sum_{q=1}^{m_l}W_{iq}^lW_{jq}^l\sigma^{(1)}(G_{q}^l(x))\sigma^{(1)}(G_{q}^l(x'))\\
    \end{aligned}
  \end{equation*}
  under $C^{0}(T\times T;\R)$ where $G^l$ is a Gaussian process in $\R^{m_l}$ whose components are i.i.d. and have mean zero and covariance kernel $K^{\RF}_l$. With weak law of large number, we obtain the finite-dimensional convergence
  \begin{equation*}
    \xk{K_{l+1,ij}^{\NT,\theta}(x_\alpha,x_\alpha')}_{\alpha\in A}\xrightarrow{p}\xk{\delta_{ij}K^{\NT}_{l+1}(x_\alpha,x_\alpha')}_{\alpha\in A}
  \end{equation*}
  for any finite set $\{(x_\alpha,x_\alpha')\in T\times T\big|\alpha\in A\}$. What we still need to prove is that for any $\delta>0$, 
  \begin{equation*}
      \norm{K_{l+1,ij}^{\NT,\theta}}_{C^{k\times k,1}(T\times T;\bbR)}\leq C
  \end{equation*}
  for some $C$ not depending on $m_1,m_2,\dots,m_l$ with probability at least $1-\delta$. Suppose that this control holds. Then, with the finite-dimensional convergence and \cref{lem:tightness_Ckk}, we obatin the conclusion \cref{eq:kernel_convergence}. Note that $T$ is convex. We have 
  \begin{equation*}
      \norm{K_{l+1,ij}^{\NT,\theta}}_{C^{k\times k,1}(T\times T;\bbR)}\leq\norm{K_{l+1,ij}^{\NT,\theta}}_{C^{(k+1)\times (k+1)}(T\times T;\bbR)}.
  \end{equation*}
  With the basic inequality $ab\leq\frac{1}{2}(a^2+b^2)$, assumption for $\sigma$ and \cref{prop:GP}, we have 
  \begin{equation*}
  \begin{aligned}
\bbE\zk{\sup_{x,x'\in T}\abs{D^\alpha\sigma(G_1^l(x))D^\beta\sigma(G_1^l(x'))}}<\infty,
  \end{aligned}
  \end{equation*}
  \begin{equation*}
      \begin{aligned}
          \bbE\zk{\sup_{x,x'\in T}\abs{{W_{11}^l}^2D^\alpha_xD^\beta_{x'}\dk{\sigma^{(1)}(G_{1}^l(x))\sigma^{(1)}(G_{1}^l(x'))K_l^\NT(x,x')}}}<\infty
      \end{aligned}
  \end{equation*}
  and
  \begin{equation*}
      \begin{aligned}
          \bbE\zk{\sup_{x,x'\in T}\abs{D^\alpha_xD^\beta_{x'}\dk{\sigma^{(1)}(G_{1}^l(x))\sigma^{(1)}(G_{1}^l(x'))K_l^\NT(x,x')}}^2}<\infty
      \end{aligned}
  \end{equation*}
  for any $\alpha,\beta$ satisfying $|\alpha|,|\beta|\leq k+1$. With the induction hypothesis and \cref{thm:init_GP}, for any $M>0$,
  \begin{equation*}
      \begin{aligned}
          &\quad\lim_{m_1,\dots,m_{l-1}\rightarrow\infty}\sup_{m_l}\bbE\zk{\sup_{x,x'\in T}\abs{\frac{\delta_{ij}}{m_l}\sum_{q=1}^{m_l}D^\alpha\sigma(z_q^l(x))D^\beta\sigma(z_q^l(x'))}\land M}\\
          &\leq\lim_{m_1,\dots,m_{l-1}\rightarrow\infty}\bbE\zk{\sup_{x,x'\in T}\abs{D^\alpha\sigma(z_1^l(x))D^\beta\sigma(z_1^l(x'))}\land M}\\
          &=\bbE\zk{\sup_{x,x'\in T}\abs{D^\alpha\sigma(G_1^l(x))D^\beta\sigma(G_1^l(x'))}\land M}\\
          &\leq \bbE\zk{\sup_{x,x'\in T}\abs{D^\alpha\sigma(G_1^l(x))D^\beta\sigma(G_1^l(x'))}}.
      \end{aligned}
  \end{equation*}
  Hence, there exists a constant $C_1$ not depending on $m_1,\dots,m_l$ and $M$ such that 
  \begin{equation*}
      \sup_{m_1,\dots,m_l}\bbE\zk{\sup_{x,x'\in T}\abs{\frac{\delta_{ij}}{m_l}\sum_{q=1}^{m_l}D^\alpha\sigma(z_q^l(x))D^\beta\sigma(z_q^l(x'))}\land M}\leq C_1.
  \end{equation*}
  and
  \begin{equation*}
  \begin{aligned}
      &\quad\bbP\xk{\sup_{x,x'\in T}\abs{\frac{\delta_{ij}}{m_l}\sum_{q=1}^{m_l}D^\alpha\sigma(z_q^l(x))D^\beta\sigma(z_q^l(x'))}>M}\\
      &\leq\frac{1}{M}\bbE\zk{\sup_{x,x'\in T}\abs{\frac{\delta_{ij}}{m_l}\sum_{q=1}^{m_l}D^\alpha\sigma(z_q^l(x))D^\beta\sigma(z_q^l(x'))}\land M}\\
      &\leq\frac{C_1}{M}.
  \end{aligned}
  \end{equation*}
  For the second term on the right of \eqref{eq:kernel}, we do the decomposition,
  \begin{equation*}
  \begin{aligned}
      &\quad\frac{1}{m_l}\sum_{q_1=1}^{m_l}\sum_{q_2=1}^{m_l}W_{iq_1}^lW_{jq_2}^l\sigma^{(1)}(z_{q_1}^l(x))\sigma^{(1)}(z_{q_2}^l(x'))K_{l,q_1q_2}^{\NT,\theta}(x,x')\\
    &=\frac{1}{m_l}\sum_{q=1}^{m_l}W_{iq}^lW_{jq}^l\sigma^{(1)}(z_{q}^l(x))\sigma^{(1)}(z_{q}^l(x'))K_{l,qq}^{\NT,\theta}(x,x')\\
    &+\frac{1}{m_l}\sum_{q_1\neq q_2}W_{iq_1}^lW_{jq_2}^l\sigma^{(1)}(z_{q_1}^l(x))\sigma^{(1)}(z_{q_2}^l(x'))K_{l,q_1q_2}^{\NT,\theta}(x,x').
  \end{aligned}
  \end{equation*}
  For the terms on the right, it is similar to demonstrate that for any $\delta>0$, there exists constants $C_2$ and $C_3$ such that 
    \begin{equation*}
        \bbP\xk{\sup_{x,x'\in T}\abs{\frac{1}{m_l}\sum_{q=1}^{m_l}W_{iq}^lW_{jq}^lD_x^\alpha D_{x'}^\beta\dk{\sigma^{(1)}(z_{q}^l(x))\sigma^{(1)}(z_{q}^l(x'))K_{l,qq}^{\NT,\theta}(x,x')}}>C_2}\leq\delta
    \end{equation*}
    and 
  \begin{equation}
  \label{eq11}
      \bbP\xk{\sup_{x,x'\in T}\dk{\frac{1}{m_l^2}\sum_{q_1\neq q_2}\xk{D^\alpha_xD^\beta_{x'}\dk{\sigma^{(1)}(z_{q_1}^l(x))\sigma^{(1)}(z_{q_2}^l(x'))K_{l,q_1q_2}^{\NT,\theta}(x,x')}}^2}> C_3}\leq \delta.
  \end{equation}
  where $C_2,C_3$ both not depneding on $m_1,m_2,\dots,m_l$.
  We define a map $F:T\times T\rightarrow\bbR^{m_l^2-m_l}$ where the components of $F(x,x')$ are given by
  \begin{equation*}
      F_{q_1q_2}(x,x')=\frac{1}{m_l}\sigma^{(1)}(z_{q_1}^l(x))\sigma^{(1)}(z_{q_2}^l(x'))K_{l,q_1q_2}^{\NT,\theta}(x,x').
  \end{equation*}
   Then, with \eqref{eq11}, for any $\delta>0$, there exsits a constant $C_4$ such that 
  \begin{equation*}
      \bbP\xk{\norm{F}_{C^{(k+1)\times(k+1)}(T\times T)}\leq C_4}\geq 1-\frac{\delta}{2}.
  \end{equation*}
  Using \cref{lem:brick} for the map $\varphi:\bbR^{m_l^2-m_l}\rightarrow\bbR$, $\varphi(x)=\sum_{q_1\neq q_2}W_{iq_1}^lW_{jq_2}^lx_{q_1q_2}$ and \cref{lem:fundamental-fact2}(for details, it is similar to the proof of \cref{lem:Lipschitz-control-net}), there also exists a constant $C_5$ such that 
  \begin{equation*}
  \begin{aligned}
      &\quad\bbP\xk{\norm{\frac{1}{m_l}\sum_{q_1\neq q_2}W_{iq_1}^lW_{jq_2}^l\sigma^{(1)}(z_{q_1}^l(x))\sigma^{(1)}(z_{q_2}^l(x'))K_{l,q_1q_2}^{\NT,\theta}(x,x')}_{C^{k\times k,1}(T\times T)}\geq C_5}\\
      &=\bbP\xk{\norm{\varphi\circ F(x,x')}_{C^{k\times k,1}(T\times T)}\geq C_5}\\
      &\geq 1-\delta.
  \end{aligned}
  \end{equation*}
\end{proof}

There is no additional obstacle to generalize \cref{thm:init_NTK} to general linear differential operators.

\begin{proposition}
  Let $T\subset\caX$ be a convex compact set. $k\geq 0$ is an integer. $\sigma$ satisfies \cref{assu:activation} for $k+2$.
  For any $\caT=\sum_{r=1}^pa_rD^{\alpha_r}$ satisfying $|\alpha_r|\leq k$ and $a_r\in C^0(T)$, fixed $l = 2,\dots,L+1$,
  fixing $m_{l}$, as $m_1,\dots,m_{l-1} \to \infty$ sequentially,
  we have
  \begin{align*}
    \caT_x \caT_{x'} K^{\NT,\theta}_{l,ij}(x,x') \xrightarrow{p} \delta_{ij} \caT_x \caT_{x'} K^{\NT}_l(x,x')
    \qq{for} i,j=1,\dots,m_l
  \end{align*}
  under $C^0(T\times T; \bbR).$
\end{proposition}

\subsection{Convergence in training}


In the following, we will use $\norm{v}_2$ for the 2-norm of a vector and $\norm{A}_2$ for the Frobinuous norm of a matrix.
Moreover, let us shorthand $\lambda_0 = \lambda_{\min}(K^\NT_{\caT}(X,X))$.
We also define $\tilde{v}^{\NTK}_t(x)$ as the NTK dynamics \cref{eq:NTK_GradientFlow} with initialization $\tilde{v}^{\NTK}_0(x) = u^{\NN}_0(x)$.


To control the training process, we first assume the following condition and establish the convergence in the training process.
We will present in \cref{lem:NTK_Continuity11} that this condition is verified for the neural network with $l=1$ and $d=1$, while extension to general cases is straightforward but very cumbersome.

\begin{condition}[Continuity of the gradient]
  \label{cond:ContinuityGradient}
  There are a function $B(m) : \R_+ \to \R_+$ satisfying $B(m)\to \infty$ as $m\to \infty$ and a monotonically increasing function $\bar{\eta}(\ep) : \R_{+} \to \R_+$ such that for any $\ep > 0$, when
  \begin{align}
    \label{eq:PerturbationW}
    \norm{W^l - W^l(0)}_2 \leq B(m) \bar{\eta}(\ep),\quad \forall l = 0,\dots,L,
  \end{align}
  we have
  \begin{align*}
    \sup_{x \in \caX} \norm{\nabla_{\theta} \caT u^{\NN}(x;\theta) - \nabla_{\theta} \caT u^{\NN}(x;\theta_0)}_{2} \leq \ep.
  \end{align*}
\end{condition}


The following theorem shows the uniform approximation (over $x \in \caX$) between the training dynamics of the neural network and the corresponding kernel regression under the physics informed loss.
The general proof idea follows the perturbation analysis in the NTK literature~\citep{arora2019_ExactComputation,arora2019fine,allen-zhu2019_ConvergenceTheory,li2024_EigenvalueDecay}, as long as we regard $\caT u^{\NN}(x;\theta)$ as a whole.

\begin{theorem}[NTK training dynamics]
  \label{thm:NTK_TrainingDynamics}
  Suppose that $\caT$ is a differential operator up to order $k$, $\sigma$ satisfies \cref{assu:activation} for $k+2$, $\lambda_0 > 0$ and \cref{cond:ContinuityGradient} holds.
  Then, it holds in probability w.r.t.\ the randomness of the initialization that
  \begin{align}
    \lim_{m \to \infty} \sup_{x \in \caX} \abs{K_{\caT,\theta}^m(x,x) - K_{\caT,\theta_0}^m(x,x)} = 0
  \end{align}
  and thus
  \begin{align}
    \lim_{m \to \infty} \sup_{t \geq 0} \sup_{x \in \caX} \abs{v^{\NN}_t(x) - \tilde{v}^{\NTK}_t(x)} = 0.
  \end{align}
\end{theorem}
\begin{remark}
  We remark here that the assumption $\lambda_0 > 0$ is not restrictive and is also a common assumption in the literature~\citep{arora2019_FinegrainedAnalysis,allen-zhu2019_ConvergenceTheory}.
  If the kernel $K^{NT}_{\caT}$ is strictly positive definite, this assumption is satisfied with probability one when the data is drawn from a continuous distribution.
\end{remark}

To prove \cref{thm:NTK_TrainingDynamics}, we first show that a perturbation bound of the weights can imply a bound on the kernel function and the empirical kernel matrix.

\begin{proposition}
  \label{prop:PertKernelMat}
  Let condition \cref{cond:ContinuityGradient} hold and let $\ep > 0$ be arbitrary.
  Then, when \cref{eq:PerturbationW} holds,
  we have
  \begin{align}
    \abs{K_{\caT,\theta}^m(x,x') - K_{\caT,\theta_0}^m(x,x')} = O(\ep),
  \end{align}
  and thus
  \begin{align*}
    \norm{K_{\caT,\theta}^m(X,X) - K_{\caT,\theta_0}^m(X,X)}_{\mathrm{op}} = O(n \ep).
  \end{align*}
\end{proposition}
\begin{proof}
  We note that
  \begin{align*}
    K_{\caT,\theta}^m(x,x') = \ang{\nabla_{\theta} \caT u^{\NN}(x;\theta), \nabla_{\theta} \caT u^{\NN}(x';\theta)},
  \end{align*}
  so the result just follows from the fact that
  \begin{align*}
    \abs{\ang{v,v'} - \ang{w,w'} } &= \abs{\ang{v-w,v'-w'} + \ang{w,v'-w'} + \ang{v-w,w'}} \\
    &\leq \norm{v-w} \norm{v'-w'} + \norm{w} \norm{v'-w'} + \norm{v-w} \norm{w'},
  \end{align*}
  where we can substitute $v = \nabla_{\theta} \caT u^{\NN}(x;\theta)$, $v' = \nabla_{\theta} \caT u^{\NN}(x';\theta)$, $w = \nabla_{\theta} \caT u^{\NN}(x;\theta_0)$, $w' = \nabla_{\theta} \caT u^{\NN}(x';\theta_0)$ and apply the conditions.
\end{proof}

\begin{proof}[Proof of \cref{thm:NTK_TrainingDynamics}]
  The proof resembles the perturbation analysis in \cite{arora2019_ExactComputation} but with some modifications.
  Using \cref{thm:init_GP} and \cref{thm:init_NTK}, there are constants $L_1,L_2$ such that the following holds at initialization with probability at least $1-\delta$ when $m$ is large enough:
  \begin{align}
    \label{eq:Proof_InitBound}
    & \norm{v(x;\theta_0)}_{C^0} \leq L_1,\\
    \label{eq:Proof_InitKernel}
    & \norm{K_{\caT,\theta_0}^m(X,X) - K^\NT_{\caT}(X,X)}_{\mathrm{op}} \leq \lambda_0/4 \\
    \label{eq:Proof_InitBoundedKernel}
    & \sup_{x \in \caX} \norm{\nabla_{\theta} \caT u^{\NN}(x;\theta_0)}_2 \leq L_2.
  \end{align}

  Let $\ep > 0$ be arbitrary.
  Using the \cref{eq:Proof_InitKernel} and also \cref{prop:PertKernelMat}, we can choose some $\eta > 0$ such that
  when $\norm{W^l(t) - W^l(0)} \leq \eta B(m)$,
  \begin{align}
    \label{eq:Proof_PerturbationAnalysis}
    \begin{aligned}
      & \norm{\nabla_{\theta} \caT u^{\NN}(x;\theta_t) - \nabla_{\theta} \caT u^{\NN}(x;\theta_0)}_{2} \leq 1, \\
      &\norm{K_{\caT,\theta_t}^m(X,X) - K_{\caT,\theta_0}^m(X,X)}_{\mathrm{op}} \leq \lambda_0/4 \\
      & \sup_{x \in \caX} \abs{K_{\caT,\theta_t}^m(x,x) - K_{\caT,\theta_0}^m(x,x)} \leq \ep.
    \end{aligned}
  \end{align}
  Combining them with \cref{eq:Proof_InitBound} and \cref{eq:Proof_InitKernel} , we have
  \begin{align}
    \label{eq:Proof_BoundeGradient}
    \norm{\nabla_{\theta} \caT u^{\NN}(x;\theta_t)}_2 \leq C_0,
  \end{align}
  for some absolute constant $C_0 > 0$, and
  \begin{align}
    \label{eq:Proof_SampleKernelEigenvalue}
    \lambda_{\min}(K_{\caT,\theta_t}^m(X,X)) \geq \lambda_0 / 2.
  \end{align}
  Now, we define
  \begin{align*}
    T_0 = \inf\dk{t \geq 0 : \norm{W^l(t) - W^l(0)} \geq \eta B(m)~\text{for some}~l}.
  \end{align*}
  Then, \cref{eq:Proof_BoundeGradient} and \cref{eq:Proof_SampleKernelEigenvalue} hold when $t \leq T_0$.

  Using the gradient flow \cref{eq:gradient-flow}, we find that
  \begin{align*}
    \dot{v}_t(X) = - \frac{1}{n} K_{\caT,\theta_t}^m(X,X) (v_t(X;\theta)-Y),
  \end{align*}
  so \cref{eq:Proof_SampleKernelEigenvalue} implies that
  \begin{align*}
    \norm{v_t(X;\theta)-Y}_2 \leq \exp(-\frac{1}{4n} \lambda_0 t) \norm{v_0(X)-Y}_2,\qq{for} t \leq T_0.
  \end{align*}
  Furthermore, we recall the gradient flow equation for $W^l$ that
  \begin{align*}
    W^l_t - W^l_0 = \int_0^t \frac{1}{n} \sum_{i=1}^n \xk{\caT u_t(x_i) - y_i} \nabla_{W^l} \caT u_t(x_i) \dd t,
  \end{align*}
  so when $t \leq T_0$, for any $l$, we have
  \begin{align*}
    \norm{W^l_t - W^l_0}_2 &\leq \frac{1}{n} \sum_{i=1}^n \int_0^t \abs{\caT u_s(x_i) - y_i} \norm{ \nabla_{W^l} \caT u_s(x_i)}_2 \dd t \\
    & \leq \frac{1}{n} \sum_{i=1}^n \sup_{s \in [0,t]} \norm{ \nabla_{W^l} \caT u_s(x_i)}_2 \int_0^t \abs{\caT u_s(x_i) - y_i} \dd t \\
    & \leq \frac{4\norm{v_0(X)-Y}_2}{\lambda_0} \sum_{i=1}^n \sup_{s \in [0,t]} \norm{ \nabla_{W^l} \caT u_s(x_i)}_2 \\
    & \leq \frac{4nC_0\norm{v_0(X)-Y}_2}{\lambda_0} \\
    & \leq \frac{4nC_0\xk{\norm{Y}_2 + \sqrt {n} L}}{\lambda_0}
  \end{align*}
  where we used the \cref{eq:Proof_BoundeGradient} in the last inequality.
  Now, as long as $m$ is large enough that
  \begin{align*}
    \eta B(m) > \frac{4nC_0\xk{\norm{Y}_2 + \sqrt {n} L}}{\lambda_0},
  \end{align*}
  an argument by contradiction shows that $T_0 = \infty$.

  Now we have shown that $\norm{W^l(t) - W^l(0)} \leq \eta B(m)$ holds for all $t \geq 0$,
  so the last inequality in \cref{eq:Proof_PerturbationAnalysis} gives
  \begin{align*}
    \sup_{t \geq 0} \sup_{x \in \caX} \abs{K_{\caT,\theta_t}^m(x,x) - K_{\caT,\theta_0}^m(x,x)} \leq \ep.
  \end{align*}
  Therefore, a standard perturbation analysis comparing the ODEs \cref{eq:gradient-flow} and \cref{eq:NTK_GradientFlow} yields the conclusion,
  see, e.g., Proof of Lemma F.1 in \cite{arora2019_ExactComputation}.

\end{proof}

In some simple cases, we can verify \cref{cond:ContinuityGradient} in a direct way.
\begin{lemma}
\label{lem:NTK_Continuity11}
    Consider neural network \eqref{eq:NN} with $l=1$ and $d=1$. Let $T\subset\caX$ be a convex compact set, $k\geq 1$ be an integer and $\sigma$ satisfy \cref{assu:activation} for $k+2$. Then, for any $\delta>0$, there exists functions $B(m)$ and $\eta(\ep)$ satisfying such that with probability at least $1-\delta$ over initialization, we have
    \begin{equation*}
        \sup_{x \in T} \norm{\nabla_{\theta} \frac{d^ku^{\NN}}{dx} (x;\theta) - \nabla_{\theta} \frac{d^ku^{\NN}}{dx} (x;\theta_0)}_{2} \leq \ep.
    \end{equation*}
    for any $W^0,W^1$ satisfying
    \begin{equation*}
        \norm{W^0-W^0(0)}_2,\norm{W^1-W^1(0)}_2\leq B(m)\eta(\ep).
    \end{equation*}
\end{lemma}
\begin{proof}
In this case, the neural network is defined as
\begin{equation*}
    u^{\NN}(x;\theta)\coloneqq z(x)=\frac{1}{\sqrt{m}}\sum_{i=1}^{m}W_i^1\sigma\xk{W_i^0 x}.
\end{equation*}
Hence,
\begin{equation*}
    z^{(k)}(x)=\frac{1}{\sqrt{m}}\sum_{i=1}^{m}W_i^1{W_i^0}^k\sigma^{(k)}\xk{W_i^0 x}
\end{equation*}
and
\begin{equation}
\label{eq21}
\begin{aligned}
    &\quad\norm{\nabla_{\theta} z^{(k)}(x;\theta) - \nabla_{\theta} z^{(k)}(x;\theta_0)}_{2}^2\\
    &\leq\frac{1}{m}\sum_{i=1}^m\xk{{W_i^0}^k\sigma^{(k)}\xk{W_i^0 x}-{W_i^0(0)}^k\sigma^{(k)}\xk{W_i^0(0) x}}^2\\
    &+\frac{k^2}{m}\sum_{i=1}^m\xk{W_i^1{W_i^0}^{k-1}\sigma^{(k)}\xk{W_i^0 x}-W_i^1(0){W_i^0(0)}^{k-1}\sigma^{(k)}\xk{W_i^0(0) x}}^2\\
    &+\frac{x^2}{m}\sum_{i=1}^m\xk{W_i^1{W_i^0}^{k}\sigma^{(k+1)}\xk{W_i^0 x}-W_i^1(0){W_i^0(0)}^{k}\sigma^{(k+1)}\xk{W_i^0(0) x}}^2.
\end{aligned}
\end{equation}
We only need to demonstrate that for any $\delta>0$, there exists $B(m)$ and $\eta(\ep)$ such that with probability at least $1-\delta$, we have
\begin{equation*}
    \sup_{x}\dk{\frac{1}{m}\sum_{i=1}^m\xk{{W_i^0}^k\sigma^{(k)}\xk{W_i^0 x}-{W_i^0(0)}^k\sigma^{(k)}\xk{W_i^0(0) x}}^2}\leq \ep
\end{equation*}
for any $\varepsilon>0$ and $W^0,W^1$ satisfying
    \begin{equation*}
        \norm{W^0-W^0(0)}_2,\norm{W^1-W^1(0)}_2\leq B(m)\eta(\ep).
    \end{equation*}
For the last two terms to the right of \eqref{eq21}, we can draw a similar conclusion using the same method since $k$ is a fixed integer and $T$ is a compact set. In fact, we first do the decomposition,
\begin{equation*}
\begin{aligned}
    &\quad\frac{1}{m}\sum_{i=1}^m\xk{{W_i^0}^k\sigma^{(k)}\xk{W_i^0 x}-{W_i^0(0)}^k\sigma^{(k)}\xk{W_i^0(0) x}}^2\\
    &\leq\frac{3}{m}\sum_{i=1}^m\xk{{W_i^0}^k-{W_i^0}(0)^k}^2\xk{\sigma^{(k)}\xk{W_i^0 x}-\sigma^{(k)}\xk{W_i^0(0) x}}^2\\
    &+\frac{3}{m}\sum_{i=1}^m\xk{{W_i^0}^k-{W_i^0}(0)^k}^2\sigma^{(k)}\xk{W_i^0(0) x}^2\\
    &+\frac{3}{m}\sum_{i=1}^m{W_i^0}(0)^{2k}\xk{\sigma^{(k)}\xk{W_i^0 x}-\sigma^{(k)}\xk{W_i^0(0) x}}^2\\
    &\leq3\dk{\frac{1}{m}\sum_{i=1}^m\xk{{W_i^0}^k-{W_i^0}(0)^k}^4\frac{1}{m}\sum_{i=1}^m\xk{\sigma^{(k)}\xk{W_i^0 x}-\sigma^{(k)}\xk{W_i^0(0) x}}^4}^\frac{1}{2}\\
    &+3\dk{\frac{1}{m}\sum_{i=1}^m\xk{{W_i^0}^k-{W_i^0}(0)^k}^4\frac{1}{m}\sum_{i=1}^m\sigma^{(k)}\xk{W_i^0(0) x}^4}^\frac{1}{2}\\
    &+3\dk{\frac{1}{m}\sum_{i=1}^m{W_i^0}(0)^{4k}\frac{1}{m}\sum_{i=1}^m\xk{\sigma^{(k)}\xk{W_i^0 x}-\sigma^{(k)}\xk{W_i^0(0) x}}^4}^\frac{1}{2}
\end{aligned}
\end{equation*}
Note that
\begin{equation*}
    \begin{aligned}
        &\quad\frac{1}{m}\sum_{i=1}^m\xk{{W_i^0}^k-{W_i^0}(0)^k}^4\\
        &\leq \frac{k^4}{m}\sum_{i=1}^m\xk{{W_i^0}-{W_i^0}(0)}^4(\abs{W_i^0(0)}+\abs{{W_i^0}-{W_i^0}(0)})^{4k-4}\\
        &\leq \frac{C_1(k)}{m}\sum_{i=1}^m\xk{{W_i^0}-{W_i^0}(0)}^4\abs{W_i^0(0)}^{4k-4}+\frac{C_1(k)}{m}\sum_{i=1}^m\xk{{W_i^0}-{W_i^0}(0)}^{4k}\\
        &\leq C_1(k)\dk{\frac{1}{m}\sum_{i=1}^m\xk{{W_i^0}-{W_i^0}(0)}^8\frac{1}{m}\sum_{i=1}^m\abs{W_i^0(0)}^{8k-8}}^\frac{1}{2}+\frac{C_1(k)}{m}\sum_{i=1}^m\xk{{W_i^0}-{W_i^0}(0)}^{4k}\\
        &\leq C_1(k)\dk{\frac{1}{m}\norm{W^0-W^0(0)}_{2}^{8}\frac{1}{m}\sum_{i=1}^m\abs{W_i^0(0)}^{8k-8}}^\frac{1}{2}+\frac{C_1(k)}{m}\norm{W^0-W^0(0)}_{2}^{4k}.
    \end{aligned}
\end{equation*}
And with \cref{assu:activation}, denoting $D\coloneqq \sup_{x\in T}|x|$, we have
\begin{equation*}
    \begin{aligned}
        &\quad\frac{1}{m}\sum_{i=1}^m\xk{\sigma^{(k)}\xk{W_i^0 x}-\sigma^{(k)}\xk{W_i^0(0) x}}^4\\
        &\leq\frac{3D^4}{m}\sum_{i=1}^m\xk{W_i^0-W_i^0(0)}^4\xk{1+D^{l_{k+1}}\xk{\abs{W_i^0-W_i^0(0)}+\abs{W_i^0(0)}}^{l_{k+1}}}^4\\
    &\leq \frac{C_2(D,l_{k+1})}{m}\sum_{i=1}^m\dk{\abs{W_i^0-W_i^0(0)}^4+\abs{W_i^0-W_i^0(0)}^{4+4l_{k+1}}+\xk{W_i^0-W_i^0(0)}^4W_i^0(0)^{4l_{k+1}}}\\
    &\leq \frac{C_2(D,l_{k+1})}{m}\xk{\norm{W^0-W^0(0)}_{2}^{4}+\norm{W^0-W^0(0)}_{2}^{4+4l_{k+1}}}\\
    &+C_2(D,l_{k+1})\dk{\frac{1}{m}\norm{W^0-W^0(0)}_{2}^{8}\frac{1}{m}\sum_{i=1}^m\abs{W_i^0(0)}^{8l_{k+1}}}^\frac{1}{2}
    \end{aligned}
\end{equation*}
With \cref{assu:activation}, the fact that $T$ is compact and any moments of $W_i^0$ is finite, we have, for any $m$,
\begin{equation*}
    \bbE\zk{\frac{1}{m}\sum_{i=1}^m\sigma^{(k)}\xk{W_i^0(0) x}^4}=\bbE\zk{\sigma^{(k)}\xk{W_1^0(0) x}^4}<\infty
\end{equation*}
and similar conclusions for any summation above not depending on $W^0$. Hence, for any $\delta>0$, there exists a constant $M>0$ not depending on $m$ such that with probability at least $1-\delta$,
\begin{equation*}
    \frac{1}{m}\sum_{i=1}^m\sigma^{(k)}\xk{W_i^0(0) x}^4,\frac{1}{m}\sum_{i=1}^m{W_i^0}(0)^{4k},\frac{1}{m}\sum_{i=1}^m{W_i^0}(0)^{8k-8},\frac{1}{m}\sum_{i=1}^m\abs{{W_i^0}(0)}^{8l_{k+1}}\leq M.
\end{equation*}
Therefore, selecting $B(m)=m^\frac{1}{d_k}$ where $d_k\coloneqq\max\dk{8,4k,4+4l_{k+1}}$ and $\eta(\ep)$ sufficiently small, we can obtain the conclusion.
\end{proof}

\begin{corollary}
  Consider the neural network \eqref{eq:NN} with $l=1$ and $d=1$. Let $T$ be a bounded closed interval in $\R$, $\caT$ is a differential operator up to order $k$, $\sigma$ satisfies \cref{assu:activation} for $k+2$ and $\lambda_0 > 0$.
  Then, in probability with respect to the randomness of the initialization, we have
  \begin{align*}
    \lim_{m \to \infty} \sup_{t \geq 0} \sup_{x \in T} \abs{v^{\NN}_t(x) - \tilde{v}^{\NTK}_t(x)} = 0.
  \end{align*}
\end{corollary}

\subsection{Impact on the Spectrum}
\label{sec:spectrum}
In the previous section, we demonstrate that the NTK related to the physics-informed loss \eqref{eq:loss} is $K_\caT^\NT=\caT_x\caT_{x'}K^\NT(x,x')$ where $K^\NT=:K_{Id}^\NT$ is the NTK of the traditional l2 loss. In this section, we present some analysis and numerical experiments to explore the impact of differential operator $\caT$ on the spectrum of integral operator with kernel $K_\caT^\NT$. Define $\caI_\caT:L^2(\caX)\rightarrow L^2(\caX)$ as the integral operator with kernel $K_\caT^\NT$, which means that for all $f\in L^2(\caX)$,
\begin{equation*}
    \caI_\caT f(x) = \int_\caX\caT_x\caT_{x'}K^\NT(x,x')f(x')dx'.
\end{equation*}
Since $\caI_\caT$ is compact and self-adjoint, it has a sequence of real eigenvalues $\{\mu_j\}_{j=1}^\infty$ tending to zero. In addition, we denote $\{\lambda_j\}_{j=1}^\infty$ as the eigenvalue of $\caI_{Id}$. Then, the following lemma shows that for a large class of $\caT$, the decay rate of $\{\mu_j\}_{j=1}^\infty$ is not faster than that of $\{\lambda_j\}_{j=1}^\infty$.
\begin{lemma}
\label{lem:spectrum}
    Suppose that $\{\mu_j\}_{j=1}^\infty>0$. Let $\caT|C_0^\infty(\caX)$ be symmetric, i.e.
    \begin{equation*}
        \ang{\caT u,v}_{L^2(\caX)} = \ang{u,\caT v}_{L^2(\caX)}
    \end{equation*}
    for all $u,v\in C_0^\infty(\caX)$ and satisfy \begin{equation*}
       C_\caT\coloneqq\sup_{v\neq 0,v\in C_0^\infty(\caX)}\frac{\norm{v}_{L^2(\caX)}}{\norm{\caT v}_{L^2(\caX)}}<\infty.  
    \end{equation*}
    Then,
    \begin{equation*}
        \sup_{j}\frac{\lambda_j}{\mu_j}\leq C_\caT^2.
    \end{equation*}
\end{lemma}
\begin{proof}
    With the definition, for all $j=1,2,\dots$,
    \begin{equation*}
        \mu_j = \min_{\dim V = j-1}\max_{v\in V^\perp}\frac{\ang{\caI_\caT v, v}}{\norm{v}^2}.
    \end{equation*}
    Note that $C_0^\infty(\caX)$ is dense in $L^2(\caX)$. Hence,
    \begin{equation*}
        \begin{aligned}
            \mu_j &= \min_{\dim V = j-1}\sup_{v\in V^\perp\cap C_0^\infty(\caX)}\frac{\ang{\caI_\caT v, v}}{\norm{v}^2}\\
            &=\min_{\dim V = j-1}\sup_{v\in V^\perp\cap C_0^\infty(\caX)}\frac{\ang{\caI_{Id} \caT v, \caT v}}{\norm{v}^2}\\
            &=:\sup_{v\in V_j^\perp\cap C_0^\infty(\caX)}\frac{\ang{\caI_{Id} \caT v, \caT v}}{\norm{v}^2}.
        \end{aligned}
    \end{equation*}
    Moreover,
    \begin{equation*}
        \begin{aligned}
            \lambda_j &= \min_{\dim V = j-1}\max_{v\in V^\perp}\frac{\ang{\caI_{Id} v, v}}{\norm{v}^2}\\
            &\leq \sup_{v\in (\caT (V_j\cap C_0^\infty(\caX)))^\perp}\frac{\ang{\caI_{Id} v, v}}{\norm{v}^2}\\
            &= \sup_{v\in (V_j\cap C_0^\infty(\caX))^\perp\cap C_0^\infty(\caX)}\frac{\ang{\caI_{Id} \caT v, \caT v}}{\norm{\caT v}^2}\\
            &=\sup_{v\in V_j^\perp\cap C_0^\infty(\caX)} \frac{\ang{\caI_{Id} \caT v, \caT v}}{\norm{\caT v}^2}.
        \end{aligned}
    \end{equation*}
    Therefore, 
    \begin{equation*}
        \lambda_j\leq \sup_{v\in V_j^\perp\cap C_0^\infty(\caX)}\frac{\ang{\caI_{Id} \caT v, \caT v}}{\norm{v}^2}\sup_{v\in V_j^\perp\cap C_0^\infty(\caX)}\frac{\norm{v}^2}{\norm{\caT v}^2}\leq C_\caT^2\mu_j.
    \end{equation*}
\end{proof}

According to the Poincar\'e inequality, the gradient operator $\nabla$ fulfills the assumptions in \cref{lem:spectrum}. These assumptions also hold for a large class of elliptic operators since their smallest eigenvalues are positive (see section 6.5.1 in \cite{evans2022partial}).

  \section{Experiments}
  \label{sec:experiments}
In this section, we present experimental results to verify our theory. Throughout the task, data $\{x_i\}_{i=1}^n$ are sampled uniformly from $[0,1]^{d}$.

\begin{figure}[htbp]
    \centering
    \begin{subfigure}{0.45\textwidth}
        \includegraphics[width=\linewidth]{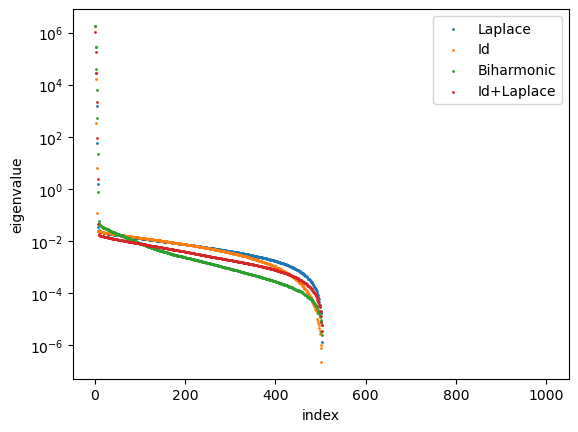}
        \caption{$d=1,\sigma(x)=Tanh(x)$}
    \end{subfigure}
    \hfill
    \begin{subfigure}{0.45\textwidth}
        \includegraphics[width=\linewidth]{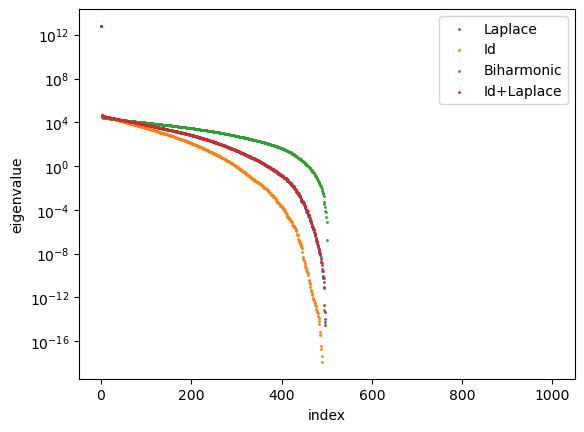}
        \caption{$d=1,\sigma(x)=ReLU(x)^6$}
    \end{subfigure}
    
    \vspace{\floatsep} 
    
    \begin{subfigure}{0.45\textwidth}
        \includegraphics[width=\linewidth]{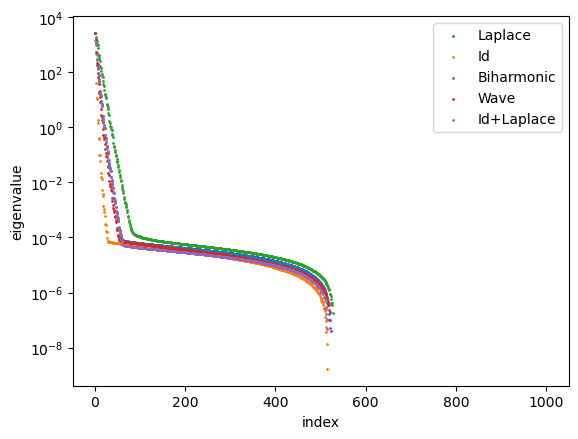}
        \caption{$d=2,\sigma(x)=Tanh(x)$}
    \end{subfigure}
    \hfill
    \begin{subfigure}{0.45\textwidth}
        \includegraphics[width=\linewidth]{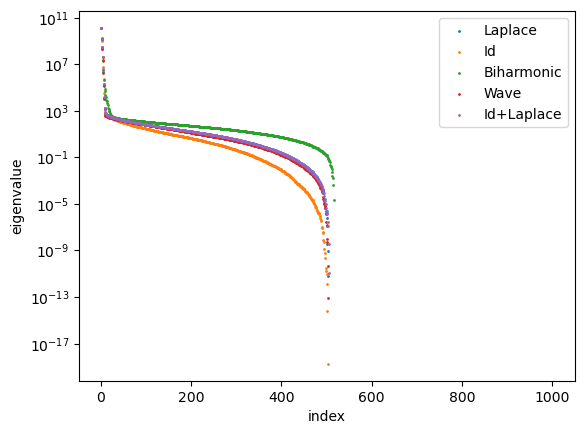}
        \caption{$d=2,\sigma(x)=ReLU(x)^6$}
    \end{subfigure}
    \caption{Eigenvalues of $K_{\caT,\theta}(X,X)$ at initialization}
    \label{figure:spectrum}
\end{figure}

In \cref{sec:spectrum}, we show that the differential operator in the loss function does not make the decay rate of the eigenvalue of the integral operator related to the NTK faster. A natural inquiry arises as to whether this phenomenon persists for the NTK matrix $K_{\caT,\theta}(X,X)$, which is closer to the neural network training dynamics~\eqref{eq:gradient-flow}. We employ the network structure described in \eqref{eq:NN} with depth $l=1$ and width $m=1024$. All parameters are initialized as independent standard normal samples. Let $n=1000$. For $d=1$, we select $\caT u(x) = u,\frac{\partial^2}{\partial x^2}u,u+\frac{\partial^2}{\partial x^2}u,\frac{\partial^4}{\partial x^4}u$. For $d=2$, we select $\caT u(x,y) = u,\Delta u,u+\Delta u,\frac{\partial^2}{\partial x^2}u-\frac{\partial^2}{\partial y^2}u,\Delta^2 u$. The activation function is Tanh or $\text{ReLU}^6$, i.e. $\sigma(x)=\max\{0,x\}^6$. The eigenvalues of $K_{\caT,\theta}(X,X)$ at initialization are shown in \cref{figure:spectrum}. Normalization is adopted to ensure that the largest eigenvalues are equal. A common phenomenon is that the higher the order of the differential operator $\caT$, the slower the decay of the eigenvalues of $K_{\caT,\theta}(X,X)$, which aligns with our theoretical predictions.
    
\begin{figure}[htbp]
    \centering
    \begin{subfigure}{0.45\textwidth}
        \includegraphics[width=\linewidth]{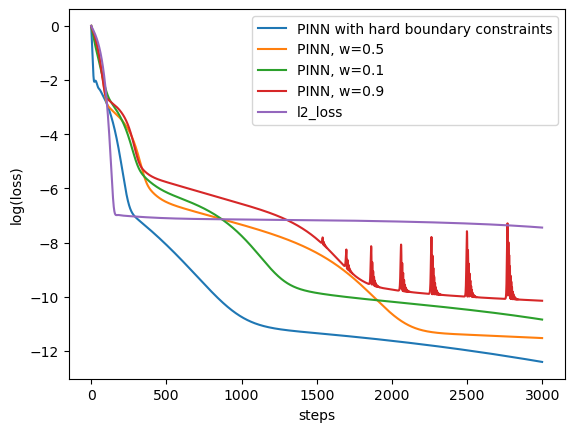}
        \caption{$a=0.5$}
    \end{subfigure}
    \hfill
    \begin{subfigure}{0.45\textwidth}
        \includegraphics[width=\linewidth]{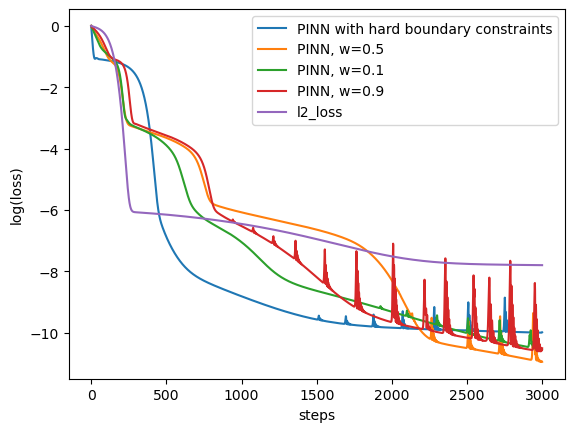}
        \caption{$a=1$}
    \end{subfigure}
    
    \vspace{\floatsep} 
    
    \begin{subfigure}{0.45\textwidth}
        \includegraphics[width=\linewidth]{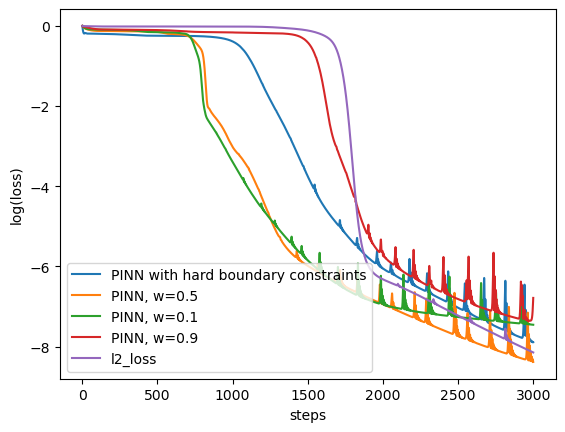}
        \caption{$a=3$}
    \end{subfigure}
    \hfill
    \begin{subfigure}{0.45\textwidth}
        \includegraphics[width=\linewidth]{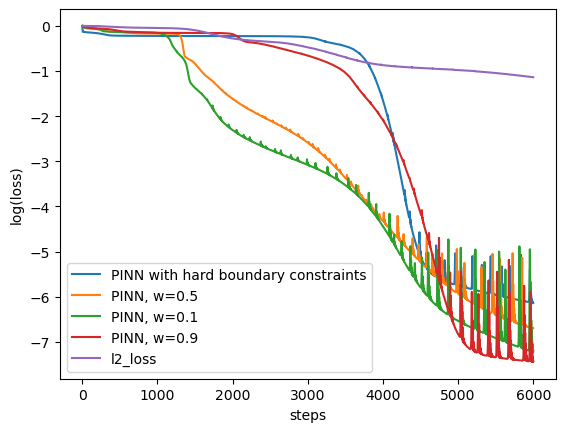}
        \caption{$a=5$}
    \end{subfigure}
    \caption{$\caL_1(u)$, $\caL_2(u)$ and $\caL_3(u)$ for different $a$ in training}
    \label{figure:training_loss}
\end{figure}

The influence of differential operators within the loss function on the actual training process is also of considerable interest. We consider to approximate $\sin(2\pi ax)$ on $[0,1]$ with positive integer $a$ and three distinct loss functions,
\begin{equation*}
\begin{aligned}
    &\caL_1(u)=\frac{1}{n}\sum_{i=1}^n\xk{D(x_i)u(x_i;\theta)-\sin(2\pi ax_i)}^2,\\
    &\caL_2(u)=\frac{1}{n}\sum_{i=1}^n\xk{-\Delta\xk{D(x_i)u(x_i;\theta)}-\sin(2\pi ax_i)}^2,\\
    &\caL_3(u;w)=\frac{1-w}{n}\sum_{i=1}^n\xk{-\Delta u(x_i;\theta)-\sin(2\pi ax_i)}^2 + w\xk{u(0;\theta)^2 + u(1;\theta)^2}
\end{aligned}
\end{equation*}
where $D(x)\coloneqq x(1-x)$ is a smooth distance function to ensure that $D(x)u(x;\theta)$ fulfills the homogeneous Dirichlet boundary condition~\cite{deng2024physical} and $\caL_3$ is the standard loss of PINNs with weight $w\in(0,1)$. In this task, we adopt a neural network setting that is more closely aligned with scenarios in practice. The network architecture in \eqref{eq:NN} is still used, but with the addition of bias terms. The parameter initialization follows the default of nn.Linear() in Pytorch. Let $l=4$, $m=512$, $n=100$ and the activation function be Tanh. We employ the Adam algorithm to train the network with learning rate 1e-5 and the nomarlized loss function $\caL_j(u(x;\theta))/\caL_j(u(x;\theta_0))$ for $j=1,2,3$. The training loss for different $a$ is shown in \cref{figure:training_loss}. For small $a$ such as $a=0.5,1$, the l2 loss $\caL_1$ decays fastest at the beginning of the training process. As $a$ increases, the decrease rate of $\caL_1$ slows down due to the constraints imposed by the spectral bias. In comparison, the loss $\caL_2$ is less affected. This corroborates that the additional differential operator in the loss function does not impose a stronger spectral bias on the neural network during training. This behavior is also observed in standard PINNs with different weights $w$.

  \section{Conclusion}
  \label{sec:conclusion}
In this paper, we develop the NTK theory for deep neural networks with physics-informed loss. We not only clarify the convergence of NTK during initialization and training, but also reveal its explicit structure. Using this structure, we prove that, in most cases, the differential operators in the loss function do not cause the neural network to face a stronger spectral bias during training. This is further supported by experiments. Therefore, if one wants to improve the performance of PINNs from the perspective of spectral bias, it may be more beneficial to focus on spectral bias caused by different terms of loss, as demonstrated in~\cite{wang2022and}. This does not mean that PINNs can better fit high-frequency functions. In fact, its training loss still decays slower when fitting a function with higher frequency, as shown in~\cref{figure:training_loss}. In instances where the solution exhibits pronounced high-frequency or multifrequency components, it is imperative to implement interventions to enhance the performance of PINNs~\cite{hedayatrasa2025k,jin2024fourier,wang2021eigenvector}. It is also important to emphasize that spectral bias is merely one aspect of understanding the limitations of PINNs, and physics-informed loss has other drawbacks, such as making the optimizing problem more ill-conditioned~\cite{krishnapriyan2021characterizing}. Hence, the addition of higher-order differential operators to address the spectral bias in the loss function is not recommended.

  \clearpage
  \bibliographystyle{plain}
  \bibliography{references}
  \clearpage
  \appendix

\section{Smoothness of the NTK}
\label{sec:smoothness}
In this section, we verify the smoothness of kernel $K^{\RF}_1$ and $K^{\NT}_1$ defined as \eqref{eq:NTK_DefRecur}. For the convenience of reader, we first recall the definition.
\begin{equation*}
  \begin{aligned}
    K^{\RF}_1(x,x') = K^{\NT}_1(x,x') = \ang{x,x'}.
  \end{aligned}
\end{equation*}
and for $l=2,\dots,L+1$,
\begin{equation*}
  \begin{aligned}
    K^{\RF}_{l}(x,x') &= \E_{(u,v)\sim N(\bm{0},\bm{B}_{l-1}(x,x'))}\left[ \sigma(u) \sigma(v) \right], \\
    K^{\NT}_{l}(x,x') &= K^{\RF}_{l}(x,x') + K^{\NT}_{l-1}(x,x') \E_{(u,v)\sim N(\bm{0},\bm{B}_{l-1}(x,x'))}\left[ \sigma^{(1)}(u) \sigma^{(1)}(v) \right],
  \end{aligned}
\end{equation*}
where the matrix $\bm{B}_{l-1}(x,x') \in \R^{2 \times 2}$ is defined as:
\[\bm{B}_{l-1}(x,x') =
\begin{pmatrix}
  K^{\RF}_{l-1}(x,x)  & K^{\RF}_{l-1}(x,x')   \\
  K^{\RF}_{l-1}(x,x') & K^{\RF}_{l-1}(x',x').
\end{pmatrix}\]


\begin{proposition}
  \label{prop:derivative-corr}
  Let $f,g$ be two functions that $f,g,f',g' \in L^2(\R,e^{-x^2/2}\dd x)$.
  Denoting
  \begin{align*}
    F(\rho) = \E_{(u,v) \sim N(\bm{0},\bm{\Sigma})} f(u) g(v), \quad
    \Sigma = \begin{pmatrix}
               1 & \rho \\ \rho & 1
    \end{pmatrix}.
  \end{align*}
  Then,
  \begin{align*}
    F'(\rho) = \E_{(u,v) \sim N(\bm{0},\bm{\Sigma})} f'(u) g'(v).
  \end{align*}
\end{proposition}
\begin{proof}
  Denote by $h_i(x)$ the Hermite polynomial basis with respect to the normal distribution.
  Let us consider the Hermite expansion of $f,g$:
  \begin{align*}
    f(u) = \sum_{i=0}^\infty \alpha_i h_i(u),\quad
    g(v) = \sum_{j=0}^\infty \beta_j h_j(v),
  \end{align*}
  and also $f',g'$:
  \begin{align*}
    f'(u) = \sum_{i=0}^\infty \alpha_i' h_i(u),\quad
    g'(v) = \sum_{j=0}^\infty \beta_j' h_j(v).
  \end{align*}
  Using the fact that $h_i'(u) = \sqrt {i} h_{i-1}(u)$, we derive
  $\alpha_{i-1}' = \sqrt {i}\alpha_i$ and $\beta_{i-1}' = \sqrt {i}\beta_i$ for $i \geq 1$.

  Now, we use the fact that $\E h_i(u) h_j(v) = \rho^i \delta_{ij}$ for $(u,v) \sim N(\bm{0},\bm{\Sigma})$ to obtain
  \begin{align*}
    \E_{(u,v)} f(u) g(v) = \sum_{i,j} \alpha_i \beta_j \E h_i(u) h_j(v) = \sum_{i=0}^\infty \alpha_i \beta_i \rho^i,
  \end{align*}
  and thus
  \begin{align*}
    \pdv{\rho} \E_{(u,v)} f(u) g(v) =  \sum_{i=0}^\infty i \alpha_i \beta_i \rho^{i-1}
    = \sum_{i=1}^\infty \alpha_{i-1}' \beta_{i-1}' \rho^{i-1} = \sum_{i=0}^\infty \alpha_i' \beta_i' \rho^i = \E_{(u,v)} f'(u) g'(v).
  \end{align*}
\end{proof}

\begin{lemma}
  \label{lem:smooth_NTK_Forward}
  Let \cref{assu:activation} hold for both $\sigma_1,\sigma_2$.
  Let us define a function $F(\bm{A})$ on positive semi-definite matrices $\bm{A} = (A_{ij})_{2 \times 2} \in \mr{PSD}(2)$ as
  \begin{equation*}
    F(\bm{A}) = \E_{(u,v)\sim N(\bm{0},\bm{A})} \sigma_1(u) \sigma_2(v).
  \end{equation*}
  Then, denoting $\caD = \dk{\bm{A} \in \mr{PSD}(2) : A_{11},A_{22} \in [c^{-1}, c] }$ for some constant $c>1$,
  we have $F(\bm{A}) \in C^k(\caD)$ and for all $\abs{\alpha} \leq k$,
  \begin{equation*}
    \abs{D^\alpha F(\bm{A})} \leq C,
  \end{equation*}
  where $C$ is a constant depending only on the constants in \cref{assu:activation} and $c$.
\end{lemma}
\begin{proof}
%

  We first prove the case for $k=1$.
  Let us consider parameterizing
  \begin{align*}
    \bm{A} =
    \begin{pmatrix}
      a^2 & \rho a b \\ \rho a b & b^2
    \end{pmatrix},
  \end{align*}
  where $\rho \in [-1,1]$ and $a,b \in [c^{-1/2}, c^{1/2}]$.
  Then,
  \begin{align*}
    F(\bm{A}) = \E_{(u,v)\sim N(\bm{0},\bm{B})} \sigma_1(a u) \sigma_2(b v),\quad
    \bm{B} =
    \begin{pmatrix}
      1 & \rho \\ \rho & 1
    \end{pmatrix}.
  \end{align*}
  Consequently,
  \begin{align*}
    \pdv{F}{a} = \E_{(u,v)\sim N(\bm{0},\bm{B})}  \xk{u\sigma_1'(a u) \sigma_2(b v)},\quad
    \pdv{F}{b} = \E_{(u,v)\sim N(\bm{0},\bm{B})}  \xk{v \sigma_1(a u) \sigma_2'(b v)}.
  \end{align*}
  Then, using \cref{assu:activation}, we have
  \begin{align*}
    \norm{\pdv{F}{a}}_{C^0}^2 \leq \zk{\E \xk{u\sigma_1'(a u)}^2}\zk{\E \xk{\sigma_2(b v)}^2}
    \leq C
  \end{align*}
  for some constant $C$.
  For $\rho$, we apply \cref{prop:derivative-corr} to get
  \begin{align*}
    \pdv{F}{\rho} = ab \E_{(u,v)\sim N(\bm{0},\bm{B})} \xk{\sigma_1'(au)} \xk{\sigma_2'(bv)}.
  \end{align*}
  Hence,
  \begin{align*}
    \abs{\pdv{F}{\rho}} \leq ab \zk{\E \xk{\sigma_1'(au)}^2}^{1/2}\zk{\E \xk{\sigma_2'(bv)}^2}^{1/2} \leq C
  \end{align*}
  for some constant $C$.

  Now, using $a = \sqrt {A_{11}}$, $b = \sqrt {A_{22}}$ and $\rho = A_{12}/\sqrt {A_{11} A_{22}}$, it is easy to see
  that the partial derivatives $\pdv{a}{A_{ij}}$, $\pdv{b}{A_{ij}}$ and $\pdv{\rho}{A_{ij}}$ are bounded by a constant depending only on $c$.
  Applying the chain rule finishes the proof for $k=1$.

  Finally, the case of general $k$ follows by induction with $\sigma_1,\sigma_2$ replaced.

\end{proof}

\begin{proof}[Proof of \cref{lem:bound-diff-NTK}]
  Using \cref{lem:smooth_NTK_Forward}, we see that the mappings
  \begin{align*}
    \bm{\Sigma} \mapsto \E_{(u,v)\sim N(\bm{0},\bm{\Sigma})}\sigma(u)\sigma(v),\quad
    \bm{\Sigma} \mapsto \E_{(u,v)\sim N(\bm{0},\bm{\Sigma})}\sigma'(u)\sigma'(v)
  \end{align*}
  belong to class $C^{k}$ and $C^{k-1}$ respectively. Hence, the result follows by the  recurrence formula and fact that compositions of $C^k$ mappings are still $C^k$.
\end{proof}
  \clearpage


\section{Auxiliary results}
The following lemma gives a sufficient condition for the weak convergence in $C^k(T)$.
\begin{lemma}[Convergence of $C^{k}$ processes]
  \label{lem:tightness_Ck}
  Let $\caX \subseteq \R^d$ be an open set.
  Let $(X^n_t)_{t \in \caX},~n\geq 1$ be random processes with $C^{k}(\caX)$ paths a.s. and $(X_t)_{t \in \caX}$ be a Gaussian field with mean zero and $C^{k\times k}$ covariance kernel. Denote by $D^\alpha$ the derivative with respect to $t$ for multi-index $\alpha$.
  Suppose that
  \begin{enumerate}
    \item For any $t_1,\dots,t_m \in T$, the finite dimensional convergence holds:
    \begin{align*}
      \xk{X^n_{t_1},\dots,X^n_{t_m}} \xrightarrow{w}
      \xk{X_{t_1},\dots,X_{t_m}}.
    \end{align*}
    \item For any $\alpha$ satisfying $|\alpha|\leq k$ and $\delta > 0$, there exists $L > 0$ such that
    \begin{align*}
        \sup_{n}\bbP\dk{\sup_{t,s\in T}\frac{\abs{D^\alpha(X_t^{n} - X_s^{n})} }{\norm{t-s}_2}> L} < \delta.
    \end{align*}
    \item For any $\alpha$ satisfying $|\alpha|\leq k$ and $\delta>0$, there exists $C>0$ such that
    \begin{align*}
        \sup_{n}\bbP\dk{\sup_{t\in T}\abs{D^\alpha X_t^{n}}>C} < \delta.
    \end{align*}
  \end{enumerate}
  Then, for any compact set $T \subset \caX$ and $\alpha$ satisfying $|\alpha|\leq k$, we have
  \begin{align*}
      D^\alpha X^n_t \xrightarrow{w} D^\alpha X_t \qq{in} C^0(T).
  \end{align*}

\end{lemma}

\begin{proof}
    With condition 2, 3 for $\alpha = \mathbf{0}$ and condition 1 (We refer to \cite[Section 23]{kallenberg2021_FoundationsModern} for more details), 
    \begin{equation*}
        X^n_t\xrightarrow{w} X_t \qq{in} C^0(T).
    \end{equation*}
    which is equivalent to
    \begin{equation*}               \liminf_{n\rightarrow\infty}\bbE\left[\varphi(X_t^n)\right]\geq\bbE\left[\varphi(X_t)\right]
    \end{equation*}
    for any bounded, nonnegative and Lipschitz continuous  $\varphi\in C(C^0(T);\bbR)$ (see, for example, in Theorem 1.3.4 of \cite{wellner2013weak}). For $\alpha$ satisfying $1\leq |\alpha|\leq k$, condition $2,3$ also derive the tightness of $\dk{D^\alpha X_t^n}_{n=1}^\infty$. Hence, we only need to show the finite dimensional convergence:
    \begin{equation}
    \label{eq4}
        \xk{D^\alpha X^n_{t_1},\dots,D^\alpha X^n_{t_l}} \xrightarrow{w}
      \xk{D^\alpha X_{t_1},\dots,D^\alpha X_{t_l}}
    \end{equation}
    for any $t_1,\dots,t_l \in T$.
    For large enough $m$, define smoothing map $J_m:C^0(T)\rightarrow C^k(T)$
    \begin{equation*}
        J_mf(x)\coloneqq \int_{\caX}R_m(|x-y|)f(y)dy
    \end{equation*}
    where $R:[0,\infty)\rightarrow[0,\infty)$ is a $C^k$ kernel with compact support set satisfying $\int_{\bbR^d}R(|y|)dy=1$. $R_m(s)\coloneqq m^dR(ms)$ is the scaled kernel. For any $\alpha$ satisfying $|\alpha|\leq k$, denote $D_m^\alpha\coloneqq D^\alpha\circ J_m$. Define $A_n\coloneqq\dk{\sup_{t,s\in T}\frac{\abs{D^\alpha(X_t^{n} - X_s^{n})} }{\norm{t-s}_2}\leq L}$. With condition 2, for any $\delta>0$, we can select $L$ large enough so that $\bbP(A_n)>1-\delta$ for all $n$. For any bounded, nonnegative and Lipschitz continuous  $\varphi\in C(\bbR^l;\bbR)$, we have
    \begin{equation}
    \label{eq2}
        \begin{aligned}
            &\quad \bbE[\abs{\varphi(D_m^\alpha X^{n}_{t_1},\dots,D_m^\alpha X^{n}_{t_l})-\varphi(D^\alpha X^{n}_{t_1},\dots,D^\alpha X^n_{t_l})}]\\
            &\leq \bbE[\abs{\varphi(D_m^\alpha X^{n}_{t_1},\dots,D_m^\alpha X^{n}_{t_l})-\varphi(D^\alpha X^{n}_{t_1},\dots,D^\alpha X^n_{t_l})}\mathbf{1}_{A_n}]\\
            &+\bbE[\abs{\varphi(D_m^\alpha X^{n}_{t_1},\dots,D_m^\alpha X^{n}_{t_l})-\varphi(D^\alpha X^{n}_{t_1},\dots,D^\alpha X^n_{t_l})}\mathbf{1}_{A_n^c}]\\
            &\leq C_1(\varphi,l)\left(\bbE[\max_{1\leq i\leq l}\abs{D_m^\alpha X^{n}_{t_i}-D^\alpha X^{n}_{t_i}}\mathbf{1}_{A_n}]+\bbP(A_n^c)\right)\\
            &\leq \frac{C_2(\varphi,R,L)}{m}+C_1(\varphi)\delta.
        \end{aligned}
    \end{equation}
    For the next part, we first consider $\alpha$ satisfying $|\alpha|=1$. Without loss of generality, we can assume that $D^\alpha=\frac{\partial}{\partial t^1}$. Since $X_t$ is a Gaussian field, $\frac{\partial X}{\partial t^1}$ exists in the sense of mean square(see, for example, Appendix 9A in \cite{edition2002probability}), i.e.
    \begin{equation*}
        \bbE\zk{\abs{\frac{X_{t+\varepsilon v}-X_t}{\varepsilon}-\frac{\partial X_t}{\partial t^1}}^2}\longrightarrow 0,\quad\text{as }\varepsilon\rightarrow 0^+
    \end{equation*}
    where $v\coloneqq(1,0,\dots,0)\in\bbR^d$. Moreover, this convergence is uniform with respect to $t$ since covariance kernel of $X_t$ belongs to $C^{1\times 1}(T)$. Therefore, we can demonstrate that
    \begin{equation}
    \label{eq1}
        \bbE\zk{\abs{\frac{\partial J_mX_t}{\partial t^1}-J_m\frac{\partial X}{\partial t^1}(t)}^2}=0.
    \end{equation}
    for any $t\in T$. In fact, 
    \begin{equation*}
        \begin{aligned}
            &\quad\frac{\partial}{\partial t^1}\int_{\caX}R_m(|t-s|)X_sds\\
            &=\lim_{\varepsilon\rightarrow 0^+}\frac{1}{\varepsilon}\dk{\int_{\caX}R_m(|t+\varepsilon v-s|)X_sds-\int_{\caX}R_m(|t-s|)X_sds}\\
            &=\lim_{\varepsilon\rightarrow 0^+}\frac{1}{\varepsilon}\dk{\int_{\bbR^d}R_m(|t+\varepsilon v-s|)X_sds-\int_{\bbR^d}R_m(|t-s|)X_sds}\\
            &=\lim_{\varepsilon\rightarrow 0^+}\int_{\bbR^d}R_m(|t-s|)\frac{X_{s+\varepsilon v}-X_s}{\varepsilon}ds.
        \end{aligned}
    \end{equation*}
    Hence, 
    \begin{equation*}
        \begin{aligned}
            &\quad\bbE\zk{\abs{\frac{\partial J_mX_t}{\partial t^1}-J_m\frac{\partial X}{\partial t^1}(t)}^2}\\
            &=\bbE\zk{\lim_{\varepsilon\rightarrow 0^+}\abs{\int_{\bbR^d}R_m(|t-s|)\xk{\frac{X_{s+\varepsilon v}-X_s}{\varepsilon}-\frac{\partial X_s}{\partial s_1}}ds}^2}\\
            &\leq \liminf_{\varepsilon\rightarrow 0^+}\bbE\zk{\abs{\int_{\bbR^d}R_m(|t-s|)\xk{\frac{X_{s+\varepsilon v}-X_s}{\varepsilon}-\frac{\partial X_s}{\partial s_1}}ds}^2}\\
            &\leq C_R\liminf_{\varepsilon\rightarrow 0^+} \int_{\bbR^d}R_m(|t-s|)\bbE\zk{\abs{\frac{X_{s+\varepsilon v}-X_s}{\varepsilon}-\frac{\partial X_s}{\partial s_1}}^2}ds\\
            &=0.
        \end{aligned}
    \end{equation*}
    Since $\frac{\partial X_t}{\partial t^1}$ is also a Gaussian field, it a.s. has uniformly continuous path . As a result, 
    \begin{equation*}
        J_m\frac{\partial X}{\partial t^1}(t)\xrightarrow{a.s.}\frac{\partial X}{\partial t^1}(t)
    \end{equation*}
    as $m\rightarrow\infty$. Note that with \cref{prop:GP},
    \begin{equation*}
        \begin{aligned}
            \sup_{m}\bbE\zk{\abs{J_m\frac{\partial X}{\partial t^1}(t)}^2}\leq C_R\bbE\zk{\sup_{t\in T}\abs{\frac{\partial X_t}{\partial t^1}}^2}<\infty
        \end{aligned}
    \end{equation*}
    which means that $\{J_m\frac{\partial X}{\partial t^1}(t) \}_{m=1}^\infty$ is uniformly integrable and
    \begin{equation*}
        J_m\frac{\partial X}{\partial t^1}(t)\xrightarrow{L^1}\frac{\partial X}{\partial t^1}(t).
    \end{equation*}
    Combining with \eqref{eq1}, we have
    \begin{equation}
    \label{eq3}
        \frac{\partial J_mX_t}{\partial t^1}\xrightarrow{L^1}\frac{\partial X}{\partial t^1}(t)
    \end{equation}
    as $m\rightarrow\infty$.
    For any fixed $m$, note that $f \mapsto \varphi\xk{\frac{\partial J_mf}{\partial t^1}(t_1),\dots,\frac{\partial J_mf}{\partial t^1}(t_l)}$ is still a bounded, nonnegative and Lipschitz continuous functional in $C(C^0(T);\bbR)$. We have
    \begin{equation*}               \liminf_{n\rightarrow\infty}\bbE\left[\varphi\xk{\frac{\partial J_mX^n}{\partial t^1}(t_1),\dots,\frac{\partial J_mX^n}{\partial t^1}(t_l)}\right]\geq\bbE\left[\varphi\xk{\frac{\partial J_mX}{\partial t^1}(t_1),\dots,\frac{\partial J_mX}{\partial t^1}(t_l)}\right]
    \end{equation*}
    Because of \eqref{eq2} and \eqref{eq3},
    \begin{equation*}               \liminf_{n\rightarrow\infty}\bbE\left[\varphi\xk{\frac{\partial X^n}{\partial t^1}(t_1),\dots,\frac{\partial X^n}{\partial t^1}(t_l)}\right]\geq\bbE\left[\varphi\xk{\frac{\partial X}{\partial t^1}(t_1),\dots,\frac{\partial X}{\partial t^1}(t_l)}\right].
    \end{equation*}
    The finite dimensional convergence \eqref{eq4} is obtained for $\alpha:|\alpha|=1$. The general situation can be processed by induction with respect to $|\alpha|$.
\end{proof}

\begin{lemma}[Convergence of $C^{k\times k}$ processes]
\label{lem:tightness_Ckk}
Let $\caX \subseteq \R^d$ be an open set.
  Let $(X^n_{t,t'})_{t,t' \in \caX},~n\geq 1$ be random processes with $C^{k\times k}(\caX\times\caX)$ paths a.s. and $(X_{t,t'})_{t,t' \in \caX}\in C^{k\times k}(\caX\times\caX)$ be a deterministic function.
  Denote by $D_t^\alpha,D_{t'}^\alpha$ the derivative with respect to $t,t'$ for multi-index $\alpha$.
  Suppose that
  \begin{enumerate}
    \item For any $(t_1,t_1'),\dots,(t_m,t_m') \in T$, the finite dimensional convergence holds:
    \begin{align*}
      \xk{X^n_{t_1,t_1'},\dots,X^n_{t_m,t_m'}} \xrightarrow{w}
      \xk{X_{t_1,t_1'},\dots,X_{t_m,t_m'}}.
    \end{align*}
    \item For any $\alpha,\beta$ satisfying $|\alpha|,|\beta|\leq k$ and $\delta > 0$, there exists $L > 0$ such that
    \begin{align*}
        \sup_{n}\bbP\dk{\sup_{t,t',s,s'\in T}\frac{\abs{D_t^\alpha D_{t'}^\beta(X_{t,t'}^{n} - X_{s,s'}^{n})} }{\norm{(t,t')-(s,s')}_2}> L} < \delta.
    \end{align*}
    \item For any $\alpha,\beta$ satisfying $|\alpha|,|\beta|\leq k$ and $\delta>0$, there exists $C>0$ such that
    \begin{align*}
        \sup_{n}\bbP\dk{\sup_{t,t'\in T}\abs{D_{t}^\alpha D_{t'}^\beta X_{t,t'}^{n}}>C} < \delta.
    \end{align*}
  \end{enumerate}
  Then, for any compact set $T \subset \caX$ and $\alpha,\beta$ satisfying $|\alpha|,|\beta|\leq k$, we have
  \begin{align*}
      D_t^\alpha D_{t'}^\beta X^n_{t,t'} \xrightarrow{w} D_t^\alpha D_{t'}^\beta X_{t,t'} \qq{in} C^0(T).
  \end{align*}
\end{lemma}

\begin{proof}
    Note that $X_{t,t'}\in C^{k\times k}(T\times T)$. For any bounded, nonnegative and Lipschitz continuous  $\varphi\in C(\bbR^l;\bbR)$, $\alpha,\beta$ satisfying $|\alpha|,|\beta|\leq k$ and $(t_1,t_1'),\dots,(t_l,t_l')\in T\times T$, we have
    \begin{equation*}
        \lim_{m\rightarrow\infty}\varphi(D_t^\alpha D_{t'}^\beta J_mX_{t_1,t_1'},\dots,D_t^\alpha D_{t'}^\beta J_mX_{t_l,t_l'})=\varphi(D_t^\alpha D_{t'}^\beta X_{t_1,t_1'},\dots,D_t^\alpha D_{t'}^\beta X_{t_l,t_l'})
    \end{equation*}
    where $J_m$ is defined in the proof of \cref{lem:tightness_Ck}. The remaining proof is the same as \cref{lem:tightness_Ck}.
\end{proof}

Let us recall our structure of neural network \cref{eq:NN},
\begin{align*}
  \begin{aligned}
    &z^{1}(x) = W^{0}x,\\
    & z^{l+1}(x) = \frac{1}{\sqrt{m_l}}W^{l}\sigma(z^{l}(x)),\quad\text{for } l=1,\dots,L
  \end{aligned}
\end{align*}
where $W^{l}\in \R^{m_{l+1} \times m_l}$. For the components,
\begin{align*}
  z^{1}_i(x) &= \sum_{j=1}^{m_0} W^{0}_{ij} x_j,\\
  z^{l+1}_i(x) &= \frac{1}{\sqrt{m_l}}\sum_{j=1}^{m_l} W^{l}_{ij} \sigma(z^{l}_j(x)).
\end{align*}
And the NNK is defined as
\begin{equation*}
    K^{\NT,\theta}_{l,ij}(x,x') = \ang{\nabla_{\theta} z^{l}_i(x), \nabla_{\theta} z^{l}_j(x')},\qq{for} i,j=1,\dots,m_l.
\end{equation*}

\begin{lemma}
    \label{lem:probability-control}
    Fix an even integer $p\geq2$. Suppose that $\mu$ is a probability measure on $\bbR$ with mean $0$ and finite higher moments. Assume also that $w=(w_1,\dots,w_{m_1})$ is a vector with i.i.d. components, each with distribution $\mu$. Fix an integer $n_0\geq 1$. Let $T_0$ be a compact set in $\bbR^{m_0}$. And $T_1\subset\bbR^{m_1}$ is the image of $T_0$ under a $C^{0,1}$ map $f$ with $\norm{f}_{C^{0,1}(T_0)}\leq \lambda$. Then, there exists a constant $C=C(T_0, p,\mu,\lambda)$ such that for all $m_1\geq1$,
\begin{equation*}
    \bbE\zk{\sup_{y\in T_1}|w\cdot y|^p}\leq C.
\end{equation*}
\end{lemma}
\begin{proof}
    For any fixed $y_0\in T_1$,
    \begin{equation*}
        \bbE\zk{\sup_{y\in T_1}|w\cdot y|^p}\leq C_1(p)\xk{\bbE\zk{|w\cdot y_0|^p}+\bbE\zk{\sup_{y\in T_1}|w\cdot (y-y_0)|^p}}.
    \end{equation*}
    With Lemma 2.9 and Lemma 2.10 in \cite{hanin2021_RandomNeural}, both two terms on the right can be bounded by a constant not depending on $m_1$.
\end{proof}

\begin{lemma}
    \label{lem:brick}
    Fix an integer $n_0\geq 1$. Let $T_0$ be a compact set in $\bbR^{m_0}$. Consider a map $\varphi:T_1\rightarrow T_2$ defined as
    \begin{equation*}
        \varphi(x)=\frac{1}{\sqrt{m_l}}\sigma(Wx)
    \end{equation*}
    where $W\in\bbR^{m_l\times m_1}$ with components drawn i.i.d. from a distribution $\mu$ with mean 0, variance 1 and finite higher moments. $T_2\subset\bbR^{m_l}$ and $T_1\subset\bbR^{m_l}$ is the image of $T_0$ under a $C^{0,1}$ map $f$ with $\norm{f}_{C^{0,1}(T_0)}\leq \lambda$. $\sigma$ satisfies \cref{assu:activation}. Then, for any $\delta>0$, there exists a positive constant $C=C(k,\lambda,\sigma,\mu,\delta)$ such that
    \begin{equation*}
        \norm{\varphi}_{C^{k,1}(T_1)}\leq C.
    \end{equation*}
    with probability at least $1-\delta$.
\end{lemma}

\begin{proof}
    For the components of $\varphi$,
    \begin{equation*}
        \varphi_i(x)=\frac{1}{\sqrt{m_l}}\sigma(W_i\cdot x).
    \end{equation*}
    Hence, for a fixed $\alpha:|\alpha|\leq k$,
    \begin{equation*}
        D^\alpha \varphi_i(x) = \frac{1}{\sqrt{m_l}}\sigma^{(|\alpha|)}(W_i\cdot x) W_i^\alpha.
    \end{equation*}
    And 
    \begin{equation*}
        \norm{D^\alpha \varphi(x)}_2^2 = \frac{1}{m_l}\sum_{i=1}^{m_l} \sigma^{(|\alpha|)}(W_i\cdot x) ^2W_i^{2\alpha}.
    \end{equation*}
    With the basic inequality $ab\leq\frac{1}{2}(a^2+b^2)$, \cref{assu:activation} and \cref{lem:probability-control}, there exists a constant $M_\alpha=M_\alpha(k,\lambda,\sigma,\mu)$ such that 
    \begin{equation*}
        \bbE \zk{\sup_{x\in T_1}\norm{D^\alpha \varphi(x)}_2^2} \leq \bbE\zk{\sup_{x\in T_1}\sigma^{(|\alpha|)}(W_i\cdot x) ^2W_i^{2\alpha}}\leq M_\alpha
    \end{equation*}
    which means that 
    \begin{equation*}
        \bbP\xk{\sup_{x\in T_1}\norm{D^\alpha \varphi(x)}_2\geq\xk{\frac{M_\alpha}{\delta}}^{\frac{1}{2}}}\leq \delta.
    \end{equation*}
    Moreover, for all $x_1,x_2\in T_1$,
    \begin{equation*}
        \begin{aligned}
            \norm{D^\alpha \varphi(x_1)-D^\alpha \varphi(x_2)}_2^2=\frac{1}{m_l}\sum_{i=1}^{m_l} \xk{\sigma^{(|\alpha|)}(W_i\cdot x_1)-\sigma^{(|\alpha|)}(W_i\cdot x_2)} ^2W_i^{2\alpha}
        \end{aligned}
    \end{equation*}
    With the same procedure in the proof of Lemma 2.11 in \cite{hanin2021_RandomNeural}, there exists a positive constant $\tilde{C}_\alpha=\tilde{C}_\alpha(k,\lambda,\sigma,\mu,\delta)$ such that
    \begin{equation*}
        \bbP\xk{\sup_{x_1,x_2\in T_1}\frac{\norm{D^\alpha \varphi(x_1)-D^\alpha \varphi(x_2)}_2}{\norm{x_1-x_2}_2}\geq\tilde{C}_\alpha}\leq \delta.
    \end{equation*}
\end{proof}

\begin{lemma}
\label{lem:Lipschitz-control-net}
    Let $\sigma$ be an activation satisfying \cref{assu:activation} and $T\subset\caX$ be a compact set in $\R^{m_0}$.
  For fixed $l = 2,\dots,L+1$ and any $\delta>0$, considering $z^{l+1}(x)$ defined as \cref{eq:NN}, there exists a positive constant $C_l=C_l(k,T,m_{l+1},\sigma,\mu,\delta)$ such that
  \begin{equation*}
 \norm{z^{l+1}(x)}_{C^{k,1}(T)}\leq C_l
  \end{equation*}
  with probability at least $1-\delta$.
\end{lemma}

\begin{proof}
    For $h=1,2,\dots,l$, define 
\begin{equation*}
    \varphi^h(x)=\frac{1}{\sqrt{m_h}}\sigma(W^{h-1}x).
\end{equation*}
and
\begin{equation*}
    \varphi^{l+1}(x)=\frac{1}{\sqrt{m_{l+1}}}W^lx
\end{equation*}
Note that
\begin{equation*}
    z^{l+1}(x)=\varphi^{l+1}\circ\varphi^{l}\circ\cdots\circ\varphi^1(x).
\end{equation*}
With \cref{lem:brick}, there exists a constant $\tilde{C}_1=\tilde{C}_1(k,T,\sigma,\mu,\delta)$ such that
\begin{equation*}
    \norm{\varphi^1(x)}_{C^{k,1}(T)}\leq \tilde{C}_1
\end{equation*}
with probability at least $1-\frac{\delta}{l+1}$. Define $A_1\coloneqq\dk{\norm{\varphi^1(x)}_{C^{k,1}(T)}\leq \tilde{C}_1}$. Then, with \cref{lem:brick}, \cref{lem:fundamental-fact} and the fact that $W^1,W^0$ are independent, there exists a constant $\tilde{C}_2=\tilde{C}_2(k,T,\sigma,\mu,\delta)$
    \begin{equation*}
    \begin{aligned}
        \bbP\xk{\norm{\varphi^2\circ\varphi^1(x)}_{C^{k,1}(T)}\leq \tilde{C}_2}&\geq\bbP\xk{\norm{\varphi^2\circ\varphi^1(x)}_{C^{k,1}(T)}\leq \tilde{C}_2,A_1}\\
        &=\bbE\zk{\mathbf{1}_{A_1}\bbP\xk{\norm{\varphi^2\circ\varphi^1(x)}_{C^{k,1}(T)}\leq \tilde{C}_2\bigg|W^0}}\\
        &\geq (1-\frac{\delta}{l+1})\bbP(A_1)\\
        &\geq 1-\frac{2\delta}{l+1}.
    \end{aligned}
    \end{equation*}
With induction, we can obtain the conclusion.
\end{proof}

\begin{lemma}
\label{lem:regularity-of-limit-kernel}
    Suppose that $\sigma$ satisfies \cref{assu:activation} for $k+1$. $T\subset\caX$ is 
 a compact set. Then, $K_l^\NT(x,x')\in C^{k\times k}(T\times T)$ and $K_l^\RF(x,x')\in C^{(k+1)\times (k+1)}(T\times T)$ where $K_l^\NT,K_l^\RF$ are defined as \cref{eq:NTK_DefRecur}.
\end{lemma}

\begin{proposition}
\label{prop:GP}
    Let $(X_t)_{t \in T}$ be a centered Gaussian process on a compact set $T \subset \R^d$ with covariance function $k(s,t) : T\times T \to \R$.
    Suppose that $k(s,t)$ is Holder-continuous, then for any $p \geq 0$ we have
    \begin{align}
        \E \sup_{t \in T} \abs{X_t}^p < \infty.
    \end{align}
\end{proposition}
\begin{proof}
    It is a standard application of Dudley’s integral, see Theorem 8.1.6 in \cite{vershynin2018_HighdimensionalProbability}.
    Since $k(s,t)$ is Holder-continuous, the canonical metric of this Gaussian process
    \begin{align*}
        d(s,t) = \zk{\E(X_s - X_t)^2}^{1/2} = \sqrt{k(s,s)+k(t,t)-2k(s,t)}
    \end{align*}
    is also Holder-continuous.
    Consequently, the covering number $\log \mathcal{N}(T,d,\ep) \lesssim \log (1/\ep)$ and the Dudley’s integral $\int_0^\infty \sqrt{\log \caN(T,d,\ep)} \dd \ep$ is finite.
    The results then follow from the tail bound
    \begin{align*}
        \bbP\dk{\sup_{s,t} \abs{X_s - X_t} \geq C \int_0^\infty \sqrt{\log \caN(T,d,\ep)} \dd \ep + u \mathrm{diam}(T)  } \leq 2\exp(-u^2).
    \end{align*}
\end{proof}

\end{document}